\documentclass[journal]{IEEEtran}

\usepackage{array,multirow,graphicx}
\usepackage{algorithm}
\usepackage{algorithmicx}
\usepackage[noend]{algpseudocode}
\usepackage{amssymb}
\usepackage{amsmath }
\usepackage{breqn}
\usepackage{xcolor}

\ifCLASSINFOpdf
\else
\fi
\begin{document}
%
\title{Continual Learning of Recurrent Neural Networks by Locally Aligning Distributed Representations}

\author{Alexander~Ororbia*, 
        Ankur~Mali, 
        C.~Lee~Giles,~\IEEEmembership{Fellow,~IEEE,}
   and~Daniel~Kifer 
\thanks{Alexander Ororbia, Computer Science, Rochester Institute of Technology, Rochester, NY 14623 USA e-mail: ago@cs.rit.edu (see https://www.cs.rit.edu/$\sim$ago/).}
\thanks{Ankur Mali, Daniel Kifer, and C. Lee Giles, Pennsylvania State University.}
\thanks{The first two authors contributed equally to this work (first author was responsible for the P-TNCN/learning algorithm/experiment design, second author was responsible for baseline models/algorithms/data). Asterisk (*) indicates corresponding author. 
}
}

%



\maketitle

\begin{abstract}
Temporal models based on recurrent neural networks have proven to be quite powerful in a wide variety of applications, including language modeling and speech processing. However, training these models often relies on back-propagation through time, which entails unfolding the network over many time steps, making the process of conducting credit assignment considerably more challenging. Furthermore, the nature of back-propagation itself does not permit the use of non-differentiable activation functions and is inherently sequential, making parallelization of the underlying training process difficult. 

Here, we propose the Parallel Temporal Neural Coding Network (P-TNCN), a biologically inspired model trained by the learning algorithm we call Local Representation Alignment. It aims to resolve the difficulties and problems that plague recurrent networks trained by back-propagation through time. The architecture requires neither unrolling in time nor the derivatives of its internal activation functions. We compare our model and learning procedure to other back-propagation through time alternatives (which also tend to be computationally expensive), including real-time recurrent learning, echo state networks, and unbiased online recurrent optimization. We show that it outperforms these on sequence modeling benchmarks such as Bouncing MNIST, a new benchmark we denote as Bouncing NotMNIST, and Penn Treebank. Notably, our approach can in some instances outperform full back-propagation through time as well as variants such as sparse attentive back-tracking. 

Significantly, the hidden unit correction phase of P-TNCN allows it to adapt to new datasets even if its synaptic weights are held fixed (zero-shot adaptation) and facilitates retention of prior generative knowledge when faced with a task sequence. We present results that show the P-TNCN's ability to conduct zero-shot adaptation and online continual sequence modeling.
\end{abstract}

%
\IEEEpeerreviewmaketitle

\section{Introduction}
\label{sec:intro}
\IEEEPARstart{L}{earning} from sequences of patterns or time series data is a crucial and challenging problem in statistical learning. Developing good models and learning procedures that can extract useful structure and form useful representations from temporal data would benefit a vast array of applications, including those in video object tracking \cite{stauffer2000learning}, human motion modeling \cite{poppe2007vision,prabhakar2010temporal}, natural language processing \cite{winograd1972understanding}, 
and even reinforcement learning, where learning a generative model of an environment can greatly aid in the act of planning \cite{walsh2010integrating}.

In recent times, temporal models based on recurrent neural networks (RNNs) have become quite prominent, achieving state-of-the-art performance in many important tasks that are sequential in nature. These tasks range from those in statistical machine translation \cite{bahdanau2014neural}, to language modeling and text processing \cite{mikolov2010recurrent,ororbia2017diff,serban2017piecewise}, to long and short-term human motion generation \cite{gopalakrishnan2018neural}, to speech recognition \cite{graves2013speech}. To train these powerful recurrent networks, \textcolor{black}{back-propagation through time (BPTT)} has long been the the primary algorithm of choice for computing parameter updates.

However, despite BPTT's popularity, this learning procedure has several important drawbacks. First, it is not a suitable choice for \emph{online} training of RNNs (where data comes in the form of a stream \cite{ororbia2017learning}). The reason is that it requires storing the input history and \emph{unfolding} the recursively defined network over this explicit chain of events, before updating parameters. 
This act of unrolling creates a large feedforward computational graph on which standard back-propagation of errors \cite{rumelhart1988backprop} (backprop) is applied. Aside from computational issues, this unrolling exacerbates one of the fundamental weaknesses of \textcolor{black}{backprop itself. Specifically, this creates an even longer global feedback pathway for error information to traverse, making the credit assignment problem} much more challenging  \cite{ororbia2018conducting,ororbia2018biologically}. A second drawback is that back-propagation is not compatible with non-differentiable activation functions (e.g., discrete \& stochastic units), thus limiting the architectural choices that users can apply to their problems.
Third, in multi-CPU/GPU setups, it is extremely difficult to parallelize the training of deep recurrent networks when using BPTT because of the strict sequential nature of backprop.

Some of the drawbacks of BPTT, such as its storage requirements, can be ameliorated by using a procedure known as \emph{truncated} back-propagation through time (TPBTT), which splits data into manageable sub-sequences.
However, this sacrifices the network's ability to capture long-term dependencies.
Given the rapidly growing use of recurrent networks in stateful problems, developing learning algorithms that can successfully resolve the limitations of BPTT is of the utmost importance. 

In this paper, we present the Parallel Temporal Neural Coding Network (P-TNCN), a model that is inspired by the Bayesian brain theory known as predictive coding \cite{rao1999predictive,panichello2013predictive}. 
Predictive coding, when applied to stateful problems, posits that many active processes in the brain, primarily those related to vision and speech, incrementally process stimuli sequences to build a dynamic, adaptable model of the world \cite{rao1997dynamic}. In essence, this is an active process \textcolor{black}{of \emph{error-correction} where the} brain first guesses what it will see and then adjusts its state in light of the actual percept sample. 
We treat the sequence modeling problem from this perspective and experimentally show that such a  process of active step-by-step prediction and state correction can allow effective incremental learning of sequences without any form of backtracking. Through the introduction of error units, we will see that the architecture of the P-TNCN allows parallelization of the various layers of deep recurrent models at both training and test time. The learning algorithm for the P-TNCN generalizes a recently proposed local learning algorithm known as Local Representation Alignment (LRA) \cite{ororbia2017learning}, which can carry out credit assignment in deep, highly nonlinear feedforward networks (e.g., with non-differentiable, stochastic, and discrete activation functions). 

The contributions we make in this article are as follows:
\begin{enumerate}
	\item We propose the P-TNCN architecture, which allows for the parallelization of training and inference.
    \item We present a learning algorithm for training the P-TNCN without unrolling, which can be viewed as a recurrent generalization of the LRA procedure. The algorithm allows the use of non-differentiable activation functions. Furthermore,  the learning procedure, combined with the P-TNCN's simpler, parameter-efficient architecture, does not incur a high computational cost/complexity (especially when compared to competing alternatives like real-time recurrent learning).
    \item We evaluate the P-TNCN on several sequence benchmarks and find that it outperforms competing online alternatives, including real-time recurrent learning and unbiased online recurrent optimization. Promisingly,  P-TNCN is competitive with full BPTT (and sparse attentive back-tracking) and can sometimes outperform it.
    \item We demonstrate that the P-TNCN can handle out-of-domain sample sequences better than models trained with full BPTT and show that it can adapt to new data even if its weights are held fixed. Furthermore, we show that the P-TNCN can effectively handle a sequence of online sequence modeling tasks with minimal forgetting.
\end{enumerate}

\section{Related Work}
\label{sec:related_work}
There has been a great deal of research in finding alternatives to back-propagation of errors, including those that are more biologically inspired \cite{lillicrap2014random,lee2015difference,scellier2017equilibrium,ororbia2018conducting,ororbia2018biologically}, but very little of it has tried to tackle the much greater challenge of learning from time-varying data points, or sequences, with some exceptions \cite{taylor2007modeling,ororbia2017learning}. Classically, a well-known online alternative to back-propagation of errors was real-time recurrent learning (RTRL, \cite{williams1989learning}), which employs forward-mode differentiation to compute gradients. However, this algorithm \textcolor{black}{scales poorly (a $4$th degree polynomial in the number of parameters)}. Some algorithms have been proposed to reduce the complexity of RTRL through (noisy) approximation, including the \emph{NoBackTrack} \cite{ollivier2015training} procedure,  \emph{Unbiased Online Recurrent Learning} \cite{tallec2017unbiased}, and the Kronecker-Factor RTRL procedure \cite{mujika2018approximating}. 
Very recently, the sparse attentive back-tracking \cite{ke2017sparse} algorithm was proposed, and while interesting in its use of an attention mechanism to choose which portions of time to propagate errors through, it still requires unfolding like BPTT.  The recurrent temporal Boltzmann machine \cite{taylor2007modeling} is an interesting Contrastive Divergence-based alternative  \cite{sutskever2009recurrent}, however, in order for the procedure to work well, BPTT is needed to effectively communicate error information across the chain of the copied restricted Boltzmann machines that operate on the sequence. 

Another way that has been proposed is to simply not learn the recurrent weights at all, as is the case for the family of \emph{reservoir computing} models, e.g., echo state networks (ESNs) \cite{jaeger2002tutorial}, Liquid State Machines (LSMs) \cite{maass2004lsm}, as well as for the algorithm known as back-propagation decorrelation \cite{schiller2005analyzing}. The primary motivation underlying reservoir computing approaches starts from the observation that the most dominant changes in the weights of an RNN occur in its output weights. This idea translates into training a large, fixed, and randomly connected RNN and modifying its output synaptic weights to learn a linear combination of the nonlinear response signals that its internal reservoir represents/contains. Since only the output weights of a reservoir computing model are adjusted, the overall training approach is far faster than even BPTT itself, but requires extensive tuning of the hyper-parameters that govern the model, particularly those that control the reservoir weight dynamics.

\subsection{Motivation: Predictive Coding}
\label{sec:pc_theory}
The neuro-cognitive motivation behind the design of our proposed model and learning procedure is grounded in the principles of predictive coding and prospective coding \cite{rainer1999prospective}.
Predictive coding theories posit that the brain is in a continuous process of creating and updating hypotheses that predict the sensory input it receives, directly influencing conscious experience \cite{panichello2013predictive}.
Models of sparse predictive coding \cite{olshausen1997sparse,rao1999predictive} embody the idea that the brain is a directed generative model where the processes of generation (top-down mechanisms) and inference (bottom-up mechanisms) are intertwined \cite{rauss2013predictive} and interact to perform a sort of iterative inference of latent variables/states. Furthermore, when nesting the ideas of predictive coding within the Kalman Filter framework \cite{rao1997dynamic}, one can create dynamic models that handle time-varying data. Many variations and implementations of predictive coding have been developed \cite{chalasani2013deep,chalasani2015contextcdn,lotter2016deep,santana2017exploiting,hwang2018robotPC,song2018fastinfpc}, some of the more recent ones merging it with back-propagation of errors as a subsequent fine-tuning step and to speed up training. 

\begin{figure*}
\centering
  \includegraphics[width=0.5\linewidth]{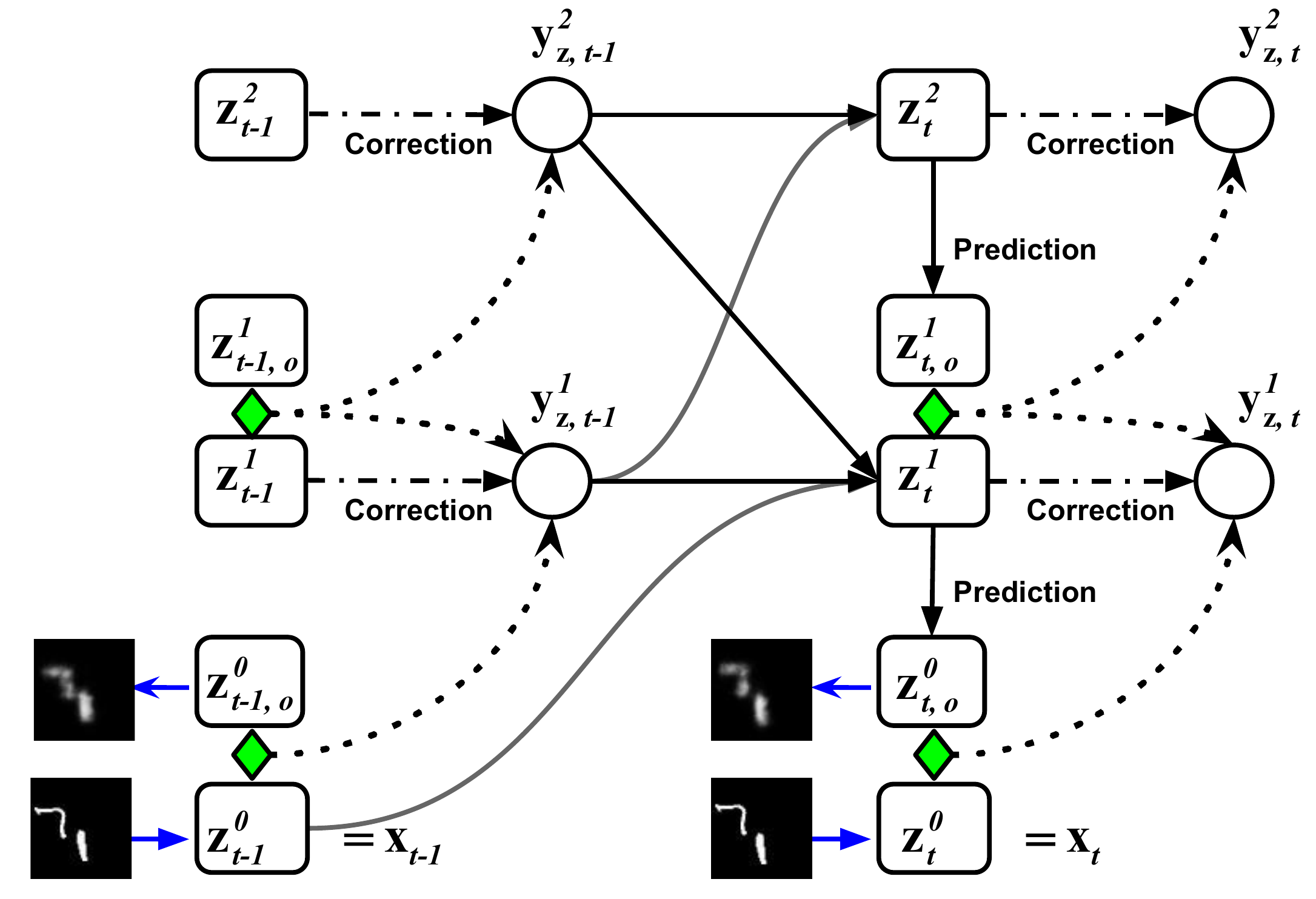}
	\caption{
	The architecture of the Parallel Temporal Neural Coding Network, shown in the act of processing data and correcting states over two time steps (during training). Dotted curves indicate pathways that error signals travel while solid lines indicate pathways that activation information travels. 	\textcolor{black}{
	The layerwise RNN temporal states are represented by $\{\mathbf{z}^1_{t},\mathbf{z}^2_{t}\}$ which generate predictions $\{\mathbf{z}^0_{t,o},\mathbf{z}^1_{t,o}\}$ of the relevant target activity regions $\{\mathbf{z}^0_{t},\mathbf{z}^1_{t}\}$. Error units (denoted $\{\mathbf{e}^0,\mathbf{e}^1\}$), graphically depicted as green diamonds, compute the mismatch between the predictions and target regions. These mismatches are transmitted across the network to adjust or ``correct'' the original temporal states $\{\mathbf{z}^1_{t},\mathbf{z}^2_{t}\}$. The adjusted values of the states are stored in the variables  $\{\mathbf{y}^1_{t,z},\mathbf{y}^2_{t,z}\}$.}
	}
\label{fig:tncn_over_time}
\vspace{-0.5cm} 
\end{figure*}

One key concept behind predictive coding that our work strongly embodies is that, for a multi-level system to work well, each layer of the neural architecture would need an error feedback mechanism to communicate its needs, i.e., activity mismatch signals, to relevant nearby regions. If the learning signals are moved closer to the layers themselves, the error connections can directly transmit the information to the right representation units. Importantly, doing this allows us to side-step the vanishing gradient problem that plagues backprop, where the internal layers of the architecture are trying to satisfy an objective that they only indirectly influence.
If we were to compare the updates from this local learning approach to backprop, the updates would still ascend/descend towards a similar objective, just not in direction of steepest ascent/descent (so long as they were within 90 degrees of the direction given by back-propagation). However, since steepest ascent/descent is a greedy form of optimization, updates from a more localized approach might lead to superior generalization results. 
Others have recently begun to investigate more local approaches to learning as well, such as kickback  \cite{balduzzi2015kickback}, which was derived specifically for regression problems and notably, decoupled neural interfaces \cite{jaderberg2016decoupled}, which tries to learn a predictive model of error gradients instead of using local information to estimate weight updates. As a result, this procedure allows layers of the underlying model to be trained independently, but in the end, relies on backprop as a subroutine.

\section{The Parallel Temporal Neural \\Coding Network}
\label{sec:tncn}
In this section, we will describe the proposed P-TNCN and its learning algorithm for computing parameter updates.

\subsection{Data and Model Architecture}
\label{sec:arch}
The P-TNCN is an architecture meant to be applied to time-varying data and variable-length sequences. One such sequence, of length $\tau$,  can be represented as $\mathbf{X} = \{  \mathbf{x}_1, \dots, \mathbf{x}_\tau \}$, where each $\mathbf{x}_i$ is a $k$-dimensional vector (for notational convenience, we  also set $\mathbf{x}_0=\mathbf{0}$).
Such a sequence could represent the words or characters that make up sentences of a document, e.g., sequence of symbols encoded as 1-of-$k$ (or one-hot) vectors, or frames of a video, e.g. sequence of flattened gray-scale pixel vectors (where the 2D pixel grid has been flattened to a 1D vector). We will examine the generative modeling of these kinds of data sequences, focusing on next-step prediction of the data point $\mathbf{x}_t$ given a history of percepts $\mathbf{x}_{<t} = \{ \mathbf{x}_0, \dots, \mathbf{x}_{t-1} \}$. In iteratively processing a sequence, at each time step, the model would be fed in a data point $\mathbf{x}_{t-1}$ and would output $\mathbf{z}^0_{t,o}$  -- its current best estimate of $\mathbf{x}_{t}$. 

The task of next-step prediction can be specifically formulated as a maximum likelihood learning problem where we are learning a graphical model of the joint distribution $p_{\Theta}(\mathbf{x}_1,\cdots,\mathbf{x}_\tau)$ decomposed as follows:
\begin{align}
p_{\Theta}(\mathbf{x}_1,\cdots,\mathbf{x}_\tau) = \prod^\tau_{t=1} p_{\Theta}(\mathbf{x}_t | \mathbf{x}_{t<}) = \prod^\tau_{t=1} p_{\Theta}(\mathbf{x}_t | \mathbf{x}_0,\cdots,\mathbf{x}_{t-1})  \label{eq:seq_prob}
\end{align}
By taking the negative of the logarithm of Equation \ref{eq:seq_prob}, we get:
\begin{align}
-\log p_\Theta(\mathbf{x}_1,\cdots,\mathbf{x}_\tau) = -\sum^\tau_{t=1} \log p_\Theta(\mathbf{x}_t | \mathbf{x}_{t<}). \label{eq:lm_loss}
\end{align}

The P-TNCN can be viewed as a set of parallel recurrent networks, coordinated by error units and simple recurrent, memory weights, working together to learn the probability distribution described above. It can be arbitrarily deep but, for the sake of illustration, we explain how it works with $2$ hidden layers of processing elements $\mathbf{z}^1_t, \mathbf{z}^2_t$, an input layer denoted by $\mathbf{z}^0_t$,  and an output layer denoted by $\mathbf{z}^0_{t,o}$. Layers $\mathbf{z}^1_t$ and $\mathbf{z}^2_t$ can be thought of as loosely-coupled \textcolor{black}{RNNs. At} time $t$, the job of  $\mathbf{z}^1_t$ (first RNN\textcolor{black}{, or RNN 1}) is to predict the next data point \textcolor{black}{$\mathbf{x}_{t+1}$} while the job of $\mathbf{z}^2_t$ (second RNN\textcolor{black}{, or RNN 2}) is to predict the next hidden state $\mathbf{z}^1_{t+1}$ of the first RNN, as illustrated in Figure \ref{fig:tncn_over_time}. We let $\mathbf{z}^1_{t+1,o}$ denote this predicted value of the state $\mathbf{z}^1_{t+1}$. After the predictions are made for the next time step (the ``prediction phase''), the next data point is observed and both RNNs correct their hidden states in an ``error-correction" phase. \textcolor{black}{For each hidden state, $\mathbf{z}^i_t$ we introduce a new variable $\mathbf{y}^i_{t,z}$ with the intuition that the network should provide better predictions if the value of the hidden state $\mathbf{z}^i_t$ was replaced by the value of $\mathbf{y}^i_{t,z}$. Thus, in the next time step $t+1$, the next hidden states $\mathbf{z}^i_{t+1}$ are computed from $\mathbf{y}^i_{t,z}$ instead of  $\mathbf{z}^i_t$.} \textcolor{black}{Thus we say that} $\mathbf{z}^1_t$ is corrected to $\mathbf{y}^1_{t,z}$ and $\mathbf{z}^2_t$ is corrected to $\mathbf{y}^2_{t,z}$. \textcolor{black}{Another way to understand the variables $\{\mathbf{y}^1_{t,z}, \mathbf{y}^2_{t,z}\}$ is to view them as: 1) latent descriptors of the P-TNCN's environment that are slightly corrected each time a new sensory vector is presented to it, and 2) target value vectors that are automatically generated to serve as teaching signals for some of the P-TNCN's synaptic weights.}

In the more general setting, with $m$ hidden layers $\mathbf{z}^1_t,\dots, \mathbf{z}^m_t$, the P-TNCN can be viewed as a set of $m$ loosely coupled RNNs. At time $t-1$, all RNNs simultaneously update their states (oblivious of each other's update) and then they predict the state of subsequent next RNN \textcolor{black}{(or predict the data, in the case of the first RNN)}. Specifically, the job of RNN 1 is to update its own state ($\mathbf{z}^1_t$) and then predict the data $\mathbf{x}_t$ at time $t$. Its input consists of (a) the data  $\mathbf{x}_{t-1}$ from the previous time step, (b) its own \emph{corrected} hidden state  $\mathbf{y}^1_{t-1,z}$ from the previous time step, and (c) the corrected hidden state $\mathbf{y}^2_{t-1,z}$ of RNN 2   from the previous time step. The job of RNN 2 is to update its own hidden state \textcolor{black}{($\mathbf{z}^2_t$)} then guess the value of the hidden state of RNN 1 at time $t$ (i.e. predict $\mathbf{z}^1_t$). Thus its inputs are the corrected hidden states of RNNs 1, 2, 3 from the previous time step (i.e. $\mathbf{y}^1_{t-1,z}, \mathbf{y}^2_{t-1,z},\mathbf{y}^3_{t-1,z}$). In general, RNN $j$ must update its own hidden state ($\mathbf{z}^j_t$) and then guess the hidden state ($\mathbf{z}^{j-1}_t$) of RNN $j-1$ using the corrected hidden states of itself and its neighboring RNNs from the previous time step (i.e., $\mathbf{y}^{j+1}_{t-1,z}$, $\mathbf{y}^{j}_{t-1,z}$, $\mathbf{y}^{j-1}_{t-1,z}$). The guess of RNN $j$ for the hidden state of RNN $j-1$ is denoted $\mathbf{z}^{j-1}_{t,o}$.
The inputs to the last RNN $m$ are $\mathbf{y}^{m-1}_{t-1,z}$ and $\mathbf{y}^{m}_{t-1,z}$ (since there is no RNN $m+1$). 

With the predictions in place, the true values of the states are observed by the neighboring RNNs. Each RNN $j$ compares its guess $\mathbf{z}^{j-1}_{t,o}$ about the next RNN to the corresponding true value $\mathbf{z}^{j-1}_t$ and then performs a self-correction, converting its own state $\mathbf{z}^j_{t}$ to the corrected state $\mathbf{y}^j_{t,z}$. These self-corrections all happen in parallel. The cycle then repeats (predict the next state, perform a self-correction, and so on).

Next, we describe the prediction and correction phases. 

\subsection{Prediction and Updating Phase}
\label{sec:prediction}
In the example of a 2 hidden state P-TNCN, the states are updated as follows.\footnote{Biases omitted for clarity. Matrices/vectors are column-major oriented.}
For efficiency, the prediction of the next step and the updates can happen in parallel. The updates are:
\begin{align}
\mathbf{a}^2_t &= V_2 \mathbf{y}^2_{t-1,z} + M_2 \mathbf{y}^1_{t-1,z} \quad\text{(pre-activation)}\\
 \mathbf{z}^2_t &= \phi^2_z( \mathbf{a}^2_t )  \label{eqn:lyr2}\\
\mathbf{a}^1_t &= U_1 \mathbf{y}^2_{t-1,z} + V_1 \mathbf{y}^1_{t-1,z} +  M_1 \mathbf{z}^0_{t-1} \quad\text{(pre-activation)}\\
\mathbf{z}^1_t &= \phi^1_z( \mathbf{a}^1_t) \label{eqn:lyr1} \\
\mathbf{z}^0_t &= \mathbf{x}_t\quad\text{(input layer)}\label{eqn:lyr0}
\end{align}
and the corresponding outputs are generated as:
\begin{align}
 \mathbf{z}^0_{t,o} = p_\Theta(\mathbf{x}_t|\mathbf{z}^1_{t}) = \phi^0_o( W_1 \mathbf{z}^1_{t}) \mbox{, and, }  \mathbf{z}^1_{t,o} &= \phi^1_o( W_2 \mathbf{z}^2_{t} ) \label{eqn:predict}
\end{align}
Note that hidden layers $1$ and $2$ simultaneously update their states using corrected states from the previous time step and use their own updated hidden states to make predictions. However, they are oblivious to each other's updated state. Thus Layer 2 makes a prediction for Layer 1's state without knowing the current state of Layer 1. Similarly, Layer 1 makes a prediction of the data without knowing the current data value.

Here the model parameters are $\Theta = \{W_1, W_2, M_1, M_2, U_1,\allowbreak V_1, V_2, E_1, E_2\}$ while $\phi^2_z, \phi^1_z, \phi^0_o$ and $\phi^1_o$ are element-wise activation functions.
Generally, $V_\ell$ is the matrix holding the recurrent synaptic weights for a neuronal layer $\ell$, $M_\ell$ is weight matrix mapping input from layer $\ell-1$ to $\ell$, $U_\ell$ contains the top-down weights that map the factors in layer $\ell+1$ to layer $\ell$, and $W_\ell$ is the matrix of prediction weights for layer $\ell$. 
%

An important property of this architecture worth emphasizing is its parallel layer-wise \textcolor{black}{execution since the hidden state at time $t$ for any layer $\ell$ is computed using} only state information from the previous time step. This means that any layers $\ell$ and $\ell^\prime$ can be placed on entirely separate computational cores.

\subsection{Error-Correction Phase}
\label{sec:error_correct}
The purpose of the error-correction phase is for each RNN to adjust its internal state based on how accurate its guesses turn out to be.
In the case of the 2-layer model, RNN 2 updates its state to $\mathbf{z}^2_t$ and produces the guess $\mathbf{z}^1_{t,o}$. Independently, RNN 1 updates its state to $\mathbf{z}^1_t$ and produces a guess $\mathbf{z}^0_{t,o}$ of the data $\mathbf{x}_t\equiv \mathbf{z}^0_t$. To reconcile the differences between guesses and true values (which are observed now that the guesses have been made), RNN 1  tries to correct its hidden state $\mathbf{z}^1_t$ with two goals in mind: 1) provide a better prediction of the data via Equation \ref{eqn:predict} and 2) to also be closer what RNN 2 predicted its state to be. Thus RNN 1 would like to locally reduce the value of the objective:
\begin{align}
\mathcal{L}^p_1(\Theta) = \frac{\beta}{2} ||\mathbf{z}^0_t - \phi^0_o(\mathbf{W}_1 \mathbf{z}^1_t)||^2 + \frac{\gamma}{2} ||\mathbf{z}^1_{t} -\mathbf{z}^1_{t,o}||^2 + \lambda |\mathbf{z}^1_t|\label{eqn:motive1}
\end{align}
by replacing $\mathbf{z}^1_t$ with a corrected value $\mathbf{y}^1_t$. Here $\beta$ is a hyper-parameter that controls the strength of the bottom-up signal, $\gamma$ is the hyper-parameter that modulates the  top-down influence, and $\lambda$ controls sparsity.

 Meanwhile, the goal of RNN 2 is to correct its own hidden state $\mathbf{z}^2_t$ to better predict the state of RNN 1 via the second half of Equation \ref{eqn:predict}, so it would like to locally reduce the value of the objective:
 \begin{align}
\mathcal{L}^p_2(\Theta) =  \frac{\beta}{2} ||\mathbf{z}^1_t - \phi^1_o(\mathbf{W}_2\mathbf{z}^2_t)||^2 + \lambda |\mathbf{z}^2_t|\label{eqn:motive2}
 \end{align}
 by replacing $\mathbf{z}^2_t$ with a corrected value $\mathbf{y}^2_t$.
 
 Now, $\mathbf{z}^1_t$ (resp., $\mathbf{z}^2_t$) is obtained by running the pre-activation $\mathbf{a}^1_t$ (resp., $\mathbf{a}^2_t$) through the activation $\phi^1_z$ (resp., $\phi^2_z$). Thus when we correct $\mathbf{z}^1_t$ to obtain $\mathbf{y}^1_t$ we want $\mathbf{y}^1_t$ to be in the \textcolor{black}{possible range of $\phi^2_z$. In essence, $\mathbf{y}^1_t$ should be ``representable'' by} the activation function (similarly, $\mathbf{y}^2_t$ should be in the range of $\phi^2_z$). Thus, instead of modifying $\mathbf{z}^1_t$ and $\mathbf{z}^2_t$ directly, we adjust their respective pre-activations $\mathbf{a}^1_t$ and $\mathbf{a}^2_t$, then run the corrected pre-activations through the respective activation functions to obtain $\mathbf{y}^1_t$ and $\mathbf{y}^2_t$. This idea of ``alignment of representations'' is one of the first ideas we borrow from LRA \cite{ororbia2018conducting}.
 
Normally, such corrections would be computed by taking a gradient descent step, differentiating Equation \ref{eqn:motive1} with respect to $\mathbf{a}^1_t$ and Equation \ref{eqn:motive2} with respect to $\mathbf{a}^2_t$. Such a choice would result in the updates:
\begin{align*}
\Delta \mathbf{a}^1_t &=A_1(B_1\beta (\mathbf{z}^0_t - \phi^0_o(\mathbf{W}_1\mathbf{z}^1_t)) - \gamma(\mathbf{z}^1_t - \mathbf{z}^1_{t,o}) -\lambda sign(\mathbf{z}^1_t))\\
\Delta \mathbf{a}^2_t &= A_2(B_2\beta(\mathbf{z}^1_t-\phi^1_o(\mathbf{W}_2\mathbf{z}^2_t)) - \lambda sign(\mathbf{z}^2_t)) 
\end{align*}
where $A_1$ is the transpose of $\frac{\partial \mathbf{z}^1_t}{\partial \mathbf{a}^1_t}$ and $B_1$ is the transpose of $\frac{\partial \phi^0_o(\mathbf{W}_1\mathbf{z}^1_t)}{\partial \mathbf{z}^1_t}$ (the matrices $A_2$ and $B_2$ are defined analogously). Now, one of the surprising results obtained from algorithms such as Feedback Alignment \cite{lillicrap2014random}, Direct Feedback Alignment \cite{nokland2016direct}, and LRA \cite{ororbia2018conducting} is that partial derivatives $B_1$ and $B_2$ involving model weights ($\mathbf{W}_1$ and $\mathbf{W}_2$) can be replaced with random matrices. Through a mechanism that is not yet completely understood \cite{lillicrap2014random,nokland2016direct}, this replacement makes training more robust and even allows networks to be trained from null initializations \cite{ororbia2018conducting}. A further improvement noted in LRA is that the derivatives of the point-wise activation functions (which in this case are $\frac{\partial \mathbf{z}^1_t}{\partial \mathbf{a}^1_t}$ and $\frac{\partial \mathbf{z}^2_t}{\partial \mathbf{a}^2_t}$ since $\mathbf{z}^1_t=\phi^1_z(\mathbf{a}^1_t)$) can be dropped as long as the activation function is monotonically non-decreasing in its input. \textcolor{black}{In the case of stochastic activation functions, the monotonic non-decreasing condition is replaced with the condition that} the output distribution for a larger input should stochastically dominate the output distribution for a smaller input. Once we drop the derivative, there is no longer any need for the activation functions to be differentiable.

With these modifications, the updates that result in ``corrected" states can be written as follows:

\begin{align} 
\mathbf{e}^0 &= -\big( \mathbf{z}^0_t - \mathbf{z}^0_{t,o} \big) \equiv -\big( \mathbf{z}^0_t - \phi^0_o(\mathbf{W}_1\mathbf{z}^1_t) \big)\label{eqn:error_units_a}\\
\mathbf{e}^1 &= -\big( \mathbf{z}^1_t - \mathbf{z}^1_{t,o} \big) \equiv -\big( \mathbf{z}^1_t - \phi^1_o(\mathbf{W}_2\mathbf{z}^2_t) \big) \label{eqn:error_units_b}\\
\mathbf{y}^2_{t,z} &=  \phi^2_z \bigg( \mathbf{a}^2_t - \Big( \beta ( E_2 \mathbf{e}^1 ) - \lambda\; \text{sign}(\mathbf{z}^2_t) \Big) \bigg) \label{eqn:y1_correct} \\
\mathbf{y}^1_{t,z} &=  \phi^1_z \bigg(\mathbf{a}^1_t - \Big( \beta ( E_1 \mathbf{e}^0 ) - \gamma \mathbf{e}^1 - \lambda\; \text{sign}(\mathbf{z}^1_t) \Big) \bigg)\label{eqn:y2_correct}
\end{align}
where $\{\lambda, \beta, \gamma\}$ are coefficients to control the strength of the sparsity penalty, the strength of the bottom-up signal, and modulation of the top-down influence. Our sparsity constraint, as indicated in Equations \ref{eqn:y1_correct} and \ref{eqn:y2_correct}, is a form of weak lateral competition, a type of activation pattern that was encouraged during iterative inference in classical sparse coding \cite{olshausen1997sparse}.

We can see that the \textcolor{black}{error units $\mathbf{e}^0$ and $\mathbf{e}^1$ play a crucial role in the P-TNCN's operation and, furthermore, that they are, in fact, the first-order derivatives of the} Gaussian log likelihood (with fixed unit variance). Note that different error units could be derived if one chose a different tactic for measuring the distance between predicted and corrected representation layers. However, the general idea is that the P-TNCN is engaged with ensuring its layer-wise representations are as close to those suggested \textcolor{black}{by the error units. It is optimizing not only on the input space, but also in the latent space, giving us some rough measure of the quality of the model's internal representations}. In some sense, this bears a loose resemblance to the \emph{bottom-up-top-down} algorithm \cite{ororbia_deep_hybrid_2015a}, which proposed a non-greedy way of learning a set of layer-wise experts.  Through the feedback mechanism and the top-down generation paths, the local learning rules of the TNCN gain some form of global coordination, which was lacking in the greedy approaches of the past \cite{bengio_greedy_2007,ororbia_deep_hybrid_2015a} when training deep belief networks and their hybrid variants.

It is important to highlight that learning and inference under this model is ideally intended to be continuous, meaning that the model simultaneously generates expectations and then corrects itself (both representations and parameters) each time a new datum from a sequence is presented. This makes the model directly suited to learning incrementally from data-streams.  
For every step of processing, the P-TNCN utilizes two types of recurrence/feedback loops to predict and correct states: 1) the model is recurrent across the temporal axis since it is stateful, which means that each processing layer depends on a vector summary of the past, and 2) the model is structurally recurrent, similar to deep Boltzmann machines and Hopfield Networks \cite{hopfield1982neural}, since error will be fed back in order to automatically correct the model's initial, or ``guessed'', representations.
As a result, the P-TNCN adapts to streams of data points through a continuous process of guess-and-check, with its guesses getting progressively better as more of a particular data sequence is presented to it. This interleaving of generation and correction fits nicely within the framework of predictive coding theory (and could be furthermore could be more concretely described as a high-level simulation of prospective coding \cite{rainer1999prospective,battaglia2004local,schutz2007prospective}), which claims that the brain is in a continuous process of creating and updating hypotheses that predict the sensory input it receives, which directly influences conscious experience \cite{panichello2013predictive}. In short, the processes of top-down generation and bottom-up inference interact to perform iterative inference over latent states \cite{rauss2013predictive}.

\subsection{Learning Algorithm: Local Representation Alignment}
\label{sec:lra}
In the 2-layer example P-TNCN, the update rules are local, which means yet further parallelization can be employed in computing changes to synaptic weights.
The rules for prediction weights $\{W_1,W_2\}$ can be derived directly from the local objectives, e.g., $\{ \mathcal{L}^p_1(\Theta), \mathcal{L}^p_2(\Theta) \}$, presented earlier. If we, furthermore, drop the derivatives of the activation functions as we did when crafting the state-update equations \textcolor{black}{(Equations \ref{eqn:y1_correct} \& \ref{eqn:y2_correct})}, we obtain the following error-driven rules:
\begin{align}
\Delta W_1 &= \mathbf{e}^0 (\mathbf{z}^1_t)^T \mbox{, and, } \Delta W_2 = \mathbf{e}^1 (\mathbf{z}^2_t)^T \label{eqn:update_W}
\end{align}
where $(\cdot)^T$ is the transpose operator.
To calculate the updates for recurrent memory weights $\{V_1,V_2\}$, the bottom-up weights $\{M_1,M_2\}$, and the top-down weights $U_1$, we must define two additional local objectives $\mathcal{L}^z_1(\Theta)$ and $\mathcal{L}^z_2(\Theta)$. These objectives dictate that, when learning, both RNN 1 and RNN 2 are to adjust the weights involved in computing their respective initial state estimates such that these estimates are closer to the targets generated by the correction phase at that time step. Specifically, these objectives are:
 \begin{align}
\mathcal{L}^z_1(\Theta) =  \frac{1}{2} ||\mathbf{y}^1_{t,z} - \mathbf{z}^1_t||^2 \mbox{, and, } \mathcal{L}^z_2(\Theta) =  \frac{1}{2} ||\mathbf{y}^2_{t,z} - \mathbf{z}^2_t||^2 \mbox{.} \label{eqn:motive3}
 \end{align}
Again, using the same approach to creating Equation \ref{eqn:update_W}, the rest of the updates, following from the objectives above, are:
\begin{align}
\Delta M_1 &= \mathbf{e}^1_{t,z} (\mathbf{z}^0_{t-1})^T \mbox{, }
\Delta V_1 = \mathbf{e}^1_{t,z} (\mathbf{z}^1_{t-1})^T \mbox{, and, } \\  
\Delta U_1 &= \mathbf{e}^1_{t,z} (\mathbf{z}^2_{t-1})^T \\
\Delta M_2 &= \mathbf{e}^2_{t,z} (\mathbf{z}^1_{t-1})^T \mbox{, and, }
\Delta V_2 = \mathbf{e}^2_{t,z} (\mathbf{z}^2_{t-1})^T 
\label{eqn:update_M_U_V}
\end{align}
where errors, $\mathbf{e}^\ell_z$, between estimated and corrected states (derived from the objectives specified in Equation \ref{eqn:motive3}) are:
\begin{align} 
\mathbf{e}^1_{t,z} &= -\big( \mathbf{y}^1_{t,z} - \mathbf{z}^1_t \big) \mbox{, and, } \mathbf{e}^2_{t,z} = -\big( \mathbf{y}^2_{t,z} - \mathbf{z}^2_t \big) \mbox{.}
\end{align}
The error synaptic weights, $\{E_1, E_2\}$, are evolved over time by applying the following proposed rule:
\begin{align}
\Delta E_1 &= (\mathbf{e}^1_{t,z} - \mathbf{e}^1_{t-1,z}) ( \mathbf{e}^0 )^T \mbox{, and, } \nonumber  \\
\Delta E_2 &= (\mathbf{e}^2_{t,z} - \mathbf{e}^2_{t-1,z}) ( \mathbf{e}^1 )^T \label{eqn:update_E}
\end{align}
where, for each set of error weights, the update depends on the temporal difference between error units at time step $t$ and $t-1$. This rule deviates from the more usual rule where the error update is proportional to the transpose of the update computed for its corresponding forward weights, as is typically done in predictive coding-based models \cite{ororbia2018biologically,rao1997dynamic}. In preliminary experiments, we found that this temporal difference rule improved generalization consistently compared to the approximate transpose rule. We speculate that this temporal difference rule is more appropriate than the approximate transpose rule since we eschew the first derivatives of the activation functions (classical predictive coding focused on linear models where the activation function was the identity). 

The rules used to adjust the P-TNCN's synaptic weights given experience follow closely to those of the LRA learning procedure \cite{ororbia2017learning,ororbia2018biologically}. LRA, in short, prescribes how a computational graph would move a set of initial layer representations towards a set of targets, which are found using a separate computational process, that better describe input/output data. \textcolor{black}{Since the earlier sections have described a process} for generating targets under the P-TNCN model, i.e., $\mathbf{y}^\ell_t$ and $\mathbf{y}^\ell_{t,z}$, all that remains to instantiate an LRA-like update procedure is to design weight adjustment rules (Equations \ref{eqn:update_W}, \ref{eqn:update_M_U_V}, \& \ref{eqn:update_E}).

Finally, as another means of introducing neuro-cognitively motivated regularization to our learned neural models we augment the update with a simple Hebbian component, motivated by early work done on LEABRA \cite{o1996biologically,oreilly1998sixprinciples} which did experiments showing that combining Hebbian learning with task-driven (error-based) learning yielded better generalization than using task-driven learning alone.
In essence, models trained with only error-based learning alone are under-constrained by the task, suffering from too much variance in the solutions ultimately found, which inhibits generalization to novel inputs. As a result, incorporating a Hebbian term\footnote{This could be viewed as a ``smarter'' form of weight decay \cite{oreilly1998sixprinciples} and is the reason why we refer to it as a neuro-cognitive regularizer.} biases the model to favor certain representations that contain not only important task-related information but also co-occurrence statistics, useful for representing an agent's environment (generatively).
This update is implemented as follows:
\begin{align}
\Delta_{Hebb} W_\ell = -\frac{\mathbf{z}^{\ell}_t (\mathbf{z}^{\ell-1}_{t-1})^T}{||\mathbf{z}^{\ell}_t (\mathbf{z}^{\ell-1}_{t-1})^T||_2} \mbox{.}
\end{align}
Note that this update is normalized by its own L2 norm, which we proposed as a simple means of preventing the unbounded weight growth that typically plagues Hebbian-like update rules. This rule is then combined with the original error-driven update as follows:
\begin{align}
\Delta W_\ell = \mathbf{e}_\ell (\mathbf{z}^{\ell-1})^T + \xi \Delta_{Hebb} W_\ell
\end{align}
where, for the left term, we multiply the error unit activities for the post-synaptic neurons by the input activities, and, for the right term, we multiply the post-synaptic neuronal activities by the input activities. Note that we introduce an additional decay factor $\xi$ to down-weight the Hebbian term to prevent it from dominating the parameter evolution process (we found that a value of $0.4$ worked fine in general).
Incorporating an unsupervised rule like a Hebbian update encourages weights of a neural model to extract general statistical structure from the data since the error units help guide the parameters towards configurations which are useful for the current task \cite{oreilly1998sixprinciples}. This same idea also appears in building successful semi-supervised learning systems, where, in a multi-objective setting, the generative criterion helps to regularize the discriminative criterion in a data-dependent way (Entropy-regularization) \cite{druck2007semi,ororbia_deep_hybrid_2015a}.
Note that while here we have shown the specific update for the prediction weights $W_{\ell}$, similar rules can be designed for the other parameters $\{M_\ell, V_\ell, U_\ell \}$ (except for error weights, which do not make use of an additional Hebbian term).

\subsection{Objective: Total Discrepancy}
\label{sec:objective}
What is the overall objective that the P-TNCN attempting to optimize? 
When we combine all of the local objectives described above, we obtain the global objective of the P-TNCN, called \emph{total discrepancy} \cite{ororbia2017learning}. This function essentially describes the level of disorder or mismatch within the neural system, and, for a 2-layer P-TNCN, this can be fully expressed as the following linear combination:
\begin{align}
\mathcal{D}(\Theta) = \mathcal{L}^p_1(\Theta) + \mathcal{L}^z_1(\Theta) + \mathcal{L}^p_2(\Theta) &+ \mathcal{L}^z_2(\Theta) \mbox{.} \label{eqn:total_discrepancy}
\end{align}
In this paper, aside from the sparsity constraints (which are Laplacian), we set the local functions above to all be a form of the Gaussian log likelihood, except for text data, where the output layer is set to be a form of the Categorical log likelihood (meaning, we modify the first term of Equation \ref{eqn:motive1}).

\section{Baseline Algorithms}
\label{sec:baselines}
We compare our proposed P-TNCN to several important baselines, with a focus on those that can train a recurrent model in an online fashion, e.g. real-time recurrent learning and a modern approximation (unbiased online recurrent optimization). In addition, we compare to a reservoir computing model, i.e., the echo state network, and to two algorithms that are based on the highly problematic and biologically implausible mechanism of unfolding, e.g., back-propagation through time and sparse attentive back-tracking. 
We implement all of the following described baselines in the codebase that supports this paper to allow for future modification as well as experimental reproducibility.
\footnote{\textcolor{black}{URL: \texttt{https://github.com/ago109/ContinualPTNCN}}}

\subsection{Real-Time Recurrent Learning}
\label{sec:rtrl}
Real-time recurrent learning (RTRL) \cite{williams1989experimental} is an online learning procedure for recurrently defined computation graphs. The aim is to optimize parameters, denoted as $\Theta$, in order to minimize a total loss for a model with a state function defined in general as:
\begin{align}
\mathbf{z}_{t+1} = F_{state}(\mathbf{x}_{t+1},\mathbf{z}_{t},\Theta) \mbox{.} \label{rtrl:eqn1}
\end{align}
RTRL computes the derivative of the states and the outputs with respect to the model weights in its forward computation while processing a sequence iteratively, i.e., without any unfolding.
For the task of next step prediction, the loss $L$ to optimize, using RTRL, is simply: 
\begin{multline}
\frac { \partial L _ { t + 1 } } { \partial \Theta } = \frac { \partial L_{t + 1}(\mathbf{y}_{t + 1}, \mathbf{y}_{t + 1}^{*} )} {\partial \mathbf{y}}  \otimes \bigg( \frac{\partial F_{\text{out}}(\mathbf{x}_{t + 1}, \mathbf{z}_{t}, \Theta)} {\partial \mathbf{z}_t} \frac{ \partial \mathbf{z}_{t} } {\partial \Theta} \\ + \frac {\partial F_ {\text{out}}(\mathbf{x}_{t + 1} , \mathbf{z}_{t} , \Theta)}{\partial \Theta} \bigg) \mbox{.} \label{rtrl:eqn2}
\end{multline}
If we differentiate Equation \ref{rtrl:eqn1} with respect to $\Theta$, we obtain:
\begin{align}
\frac{\partial \mathbf{z}_t+1}{\partial \Theta} = \frac{\partial F_{\text {state}}(\mathbf{x}_{t+1}, \mathbf{z}_{t}, \Theta)}{\partial \Theta} + \frac{\partial F_{\text {state}}(\mathbf{x}_{t+1}, \mathbf{z}_{t}, \Theta)}{\partial \mathbf{z}_t} \otimes \frac{\partial \mathbf{z}_t}{\partial \Theta} \label{rtrl:eqn3}
\end{align}
at each time we compute $\frac{\partial \mathbf{z}_t}{\partial \Theta}$ based on $\frac{\partial \mathbf{z}_t-1}{\partial \Theta}$ and then use these values to directly compute $\frac{\partial \mathbf{z}_t+1}{\partial \Theta}$. 

The above, in effect, is how RTRL calculates its gradients without resorting to backward transfer or unfolding of internal recurrence relations. Since the shape/size of $\frac{\partial \mathbf{z}_t}{\partial \Theta}$ is equal to $|z| \times |\Theta|$, for standard recurrent neural networks with $n$ hidden units, this calculation scales as ${n}^4$ (time complexity \cite{williams1995gradient}). This high complexity makes RTRL highly impractical for training wider and deeper recurrent neural models.

\subsection{Unbiased Online Recurrent Optimization}
\label{sec:uoro}
Unbiased Online Recurrent Optimization (UORO) \cite{tallec2017unbiased} uses a rank-one trick to approximate the operations involved in RTRL's gradient computation, which helps to reduce the overall complexity of the learning procedure. For instance, for any given unbiased estimation of $\frac{\partial \mathbf{z}_t}{\partial \Theta}$, we can form a stochastic matrix $\tilde{Z}_t$ such that $\mathbb{E}(\tilde{Z}_{t}) = \frac{\partial \mathbf{z}_t}{\partial \Theta}$. 
Since equation \ref{rtrl:eqn2} and \ref{rtrl:eqn3} are affine in $\frac{\partial \mathbf{z}_t}{\partial \theta}$ , unbiasedness is preserved due to the linearlity of the expectation/mean. 
We compute the value of $\tilde{Z}_t$ and plug it into \ref{rtrl:eqn2} and \ref{rtrl:eqn3} to calculate the value for $\frac{\partial \mathbf{L}_t+1}{\partial \Theta}$ and $\frac{\partial \mathbf{z}_t+1}{\partial \Theta}$.
For a rank-one, unbiased approximation, $\tilde{Z}_t = \tilde{z}_t \otimes \tilde{\Theta}_t$ at time step $t$ .To calculate $\hat{Z}_t+1$ at $t+1$ we can plug in $\tilde{Z}_t$ into \ref{rtrl:eqn3}. However, mathematically, the above equation is not yet a rank-one approximation of RTRL. 

In order to obtain a proper rank-one approximation, we must make use an efficient approximation technique proposed in \cite{ollivier2015training} where we rewrite the above equation as:
\begin{multline}
\tilde{Z}_{t+1} = \bigg( \rho_{0}\frac{\partial F_{\text{state}}(\mathbf{x}_{t+1},\mathbf{z}_{t},\theta)}{\partial \mathbf{z}} \tilde{\mathbf{z}}_t + \rho_{1}\nu \bigg) \\ \otimes \bigg( \frac{\tilde{\theta}_t}{\rho_{0}} + \frac{(\nu)^T }{\rho_{1}}\frac{\partial F_{\text{state}}(\mathbf{x}_{t+1},\mathbf{z}_{t},\theta)}{\partial \theta} \bigg)
\end{multline}
where $\nu$ is a vector of independent, random signs. $\rho$ contains $k$ positive numbers and the rank one trick can be applied for any $\rho$.  In UORO, $\rho_0$ and $\rho_1$ are meant to control the variance of the derivative approximations. In practice, we define $\rho_0$ as:
\begin{align}
\rho_{0} = \sqrt[]{\frac{\| \tilde{\theta}_t \|}{\| \frac{\partial F_{state}(\mathbf{x}_{t+1},\mathbf{z}_{t},\theta)}{\partial \mathbf{z}}\tilde{\mathbf{z}}\|}}
\end{align}
and $\rho_1$ is defined to be:
\begin{align}
\rho_{1} = \sqrt[]{\frac{\| (\nu)^T \frac{\partial F_{state}(\mathbf{x}_{t+1},\mathbf{z}_{t},\theta)}{\theta} \|}{\|\nu \|}}\mbox{.}
\end{align}
Note that initially, $\tilde{\mathbf{z}}_0 = 0$ and $\tilde{\Theta}_0 = 0$, which, as argued in \cite{tallec2017unbiased}, yields unbiased estimates at time $t = 0$. Given the construction of the UORO procedure all subsequent estimates can be shown, by induction, to be unbiased as well.

\subsection{Echo State Networks}
\label{sec:esn}
The echo state network (ESN) \cite{jaeger2002tutorial} is a special type of recurrent neural network that has also been argued to be biologically plausible. The ESN only allows a small fraction of its recurrent weights to be active for any given hidden unit and utilizes \textcolor{black}{``dynamic reservoirs''} in the hidden layer to preserve model capacity. Since the structure of reservoirs are complex, the ESN has the ability to model complex dynamical systems. 

An ESN has three weight matrices, i.e., input weight matrix $\mathbf{W}_{x}$, reservoir weight matrix $\mathbf{W}_{r}$, and output weight matrix $\mathbf{W}_{y}$. An ESN RNN consists of leaky-integrate (discrete-time) continuous-valued units. The update equations are:
\textcolor{black}{
\begin{align}
\tilde{\mathbf{z}}_r(t) &= tanh(\mathbf{W}_{x}[1:x(t)]+\mathbf{W}_{r}(t-1)) \label{esn_eqn1} \\
\mathbf{z}(t) &= (1 - \alpha)\mathbf{z}(t-1) + \alpha \tilde{\mathbf{z}}_r(t)  \label{esn_eqn2}
\end{align}
}
where the function $tanh(\cdot)$ is applied element-wise, $[1 : x(t)]$ represents the set of reservoir weights applied to each input in a sequence from $[1,t]$.  $\mathbf{z}(t) \in \mathbb{R}^{N_{\mathbf{z}}}$ represents a vector of reservoir activities and  $\tilde{\mathbf{z}(t)}\in \mathbb{R}^{N_{\mathbf{z}}}$ gives the updates at step $t$. $\mathbf{W}_{\mathbf{z}} \in \mathbb{R}^{N_{\mathbf{z}} \times (1 + {N_{\mathbf{x}}}})$ is the input weight matrix and $\mathbf{W}_{r} \in \mathbb{R}^{N_{\mathbf{z}} \times N_{\mathbf{z}}}$ are the recurrent weights. $\alpha \in (0,1)$ is the leak rate. 
\vspace{-0.7cm}

\subsection{Baseline Procedures Based on Unfolding}
\label{sec:unfold_baselines}
Other baseline learning procedures that we will compare to include BPTT \cite{werbos1988generalization,werbos1990backpropagation} and its truncated version, \textcolor{black}{truncated BPTT (or TBPTT)}. Furthermore, we compare to a recently proposed variation of BPTT/TBPTT called sparse attentive backtracking procedure \cite{ke2017sparse}, or SAB, which, in effect, learns an attention mechanism that is used to selectively back-propagate error gradients through hidden states (of the unrolled network) that have been assigned high attention weights.
\vspace{-0.5cm}

\textcolor{black}{
\subsection{On Algorithmic Complexity}
\label{sec:complexity}
In terms of algorithmic spatial complexity, the P-TNCN (trained via LRA) is desirably the same as UORO and BPTT for both offline and online learning scenarios. In terms of temporal complexity, for offline learning, the P-TNCN is as fast as UORO and BPTT (which are the same). For online learning, the P-TNCN is as fast as UORO, making it faster than the BPTT/TBPTT and far faster than RTRL. This makes the P-TNCN attractive as a potential model for learning temporal streams online, given that it is no more resource or compute-intensive as one of the current best online algorithms, UORO, and, as we will show next, it outperforms UORO consistently across all modeling tasks we investigate in this article.
In the appendix, we provide details of the complexity analysis.
}

\section{Experiments}
\label{sec:results}

\subsection{Datasets}
\label{sec:data}

\subsubsection{Bouncing MNIST} This task was designed based on the description presented in \cite{srivastava2015video}. In this sequence dataset, each video is set to be 20 frames long and consists of two digits moving or ``bouncing'' around a \textcolor{black}{$64\times64$} patch. The digits within each sequence are chosen randomly from the original MNIST dataset and placed at random initial locations within the overall patch. Each digit is assigned a velocity\footnote{The direction was sampled randomly from the unit circle and the magnitude was sampled uniformly over a fixed range.} and simply bounces off edges of the overall frame, overlapping if the digits are at the same location. This task is quite challenging due to the occlusions and the dynamics of bouncing off the walls. We generate a small initial training sample of $10,000$ sequences and report the cross entropy on the exact test-set\footnote{URL: http://www.cs.toronto.edu/$\sim$nitish/unsupervised\_video/} of \cite{srivastava2015video}.

\subsubsection{Bouncing NotMNIST} \textcolor{black}{The (static) NotMNIST} database\footnote{URL: http://yaroslavvb.blogspot.com/2011/09/notmnist-dataset.html} is a more difficult variation of MNIST created by replacing the digits with characters of varying fonts/glyphs (letters A-J). We extend this data to create a new sequence benchmark. The properties of the video samples, e.g., dimensions, sequence length, etc., are made identical to those of Bouncing MNIST. A test-set is also created (same size as that of Bouncing MNIST).

\subsubsection{Bouncing Fashion MNIST} This dataset \cite{xiao2017fashion}
contains \textcolor{black}{$28\times28$} grey-scale images of 10 classes of clothing items instead of digits or characters. Properties and preprocessing of the generated sequences are kept the same MNIST and NotMNIST. A test set of 10000 samples was used for the one-shot/zero-shot experiments described later.

\subsubsection{Penn Treebank} The Penn Treebank corpus \cite{marcus1993building} is often used to benchmark both word and character-level models via perplexity or bits-per-character.\footnote{To be directly comparable with previously reported results, we make use of the specific pre-processed train/valid/test splits found at http://www.fit.vutbr.cz/$\sim$imikolov/rnnlm/.} The corpus contains 42,068 sentences (971,657 tokens, average token-length of about 4.727 characters) of varying length (the range is from 3 to 84 tokens, at the word-level).
The vocabulary for the character-level models includes 49 unique symbols (including one for spaces). 
For the character-level models, we report the standard bits-per-character (BPC), which is a function of log likelihood. 

\begin{table*}[t]
\begin{center}
\caption{Performance of the P-TNCN versus LSTMs trained via other approaches on the Bouncing MNIST (cross-entropy), \textcolor{black}{Bouncing NotMNIST} (cross-entropy), and Penn Treebank (bits-per-character, BPC) next-step prediction problems. Note that ``\texttt{impl.}'' indicates our implementation (since the baseline would not have been previously applied to the current problem/dataset). 
}
\label{results:metrics}
\begin{tabular}{c lll||ll } 
 \hline 
 & \bf Bouncing MNIST &  &  & \bf Bouncing NotMNIST & \\
 \hline 
 & \bf Model & \bf Test CE & \bf Test SE & \bf Test CE & \bf Test SE\\
 \hline
 \hline
 \parbox[t]{0.15mm}{\multirow{6}{*}{\rotatebox[origin=c]{90}{w/ unfold}}} & LSTM-FP \cite{srivastava2015video} (BPTT) & $350.2$ & $--$ & $--$ & $--$\\ 
 & LSTM-CFP \cite{srivastava2015video} (BPTT) & $341.2$ & $--$ & $--$ & $--$\\ 
 & LSTM, BPTT (impl.) & $375.42$ & $85.27$ & $787.51$ & $256.66$\\ 
 & LSTM, SAB (impl.) & $379.3$ & $86.79$ & $787.59$ & $256.89$\\
 & \textcolor{black}{GRU, BPTT (impl.)} & \textcolor{black}{$375.0$} & \textcolor{black}{$85.18$} & \textcolor{black}{$788.00$} & \textcolor{black}{$257.01$}\\
 & RNN, BPTT (impl.) & $391.4$ & $90.14$ & $795.12$ & $269.29$\\
 & RNN, SAB (impl.) & $392.7$ & $90.22$ & $794.21$ & $265.21$\\
 \hline
 \parbox[t]{0.15mm}{\multirow{4}{*}{\rotatebox[origin=c]{90}{no unfold}}} & ESN (impl.) & $489.2$ & $99.86$ & $812.43$ & $305.57$ \\ 
 & LSTM, UORO (impl.) & $386.7$ & $89.21$ & $789.48$ & $259.10$ \\ 
 & LSTM, RTRL (impl.) & $361.2$ & $85.89$ & $778.29$ & $222.08$ \\ 
 & P-TNCN (ours) & $\mathbf{338.79}$ & $\mathbf{79.67}$ & $\mathbf{713.67}$ & $\mathbf{176.73}$ \\
 \hline
\end{tabular}
\begin{tabular}{ llll } 
 \hline 
 & \bf Penn Treebank &  &  \\
 \hline 
 & \bf Model & \bf Valid BPC & \bf Test BPC \\
 \hline
 \hline
 \parbox[t]{0.15mm}{\multirow{5}{*}{\rotatebox[origin=c]{90}{w/ unfold}}} & LSTM, TBPTT-25 \cite{mujika2018approximating} & $1.61$ & $1.56$ \\ 
 & LSTM, BPTT \cite{ke2017sparse} & $1.48$ & $1.38$ \\ 
 & LSTM, SAB \cite{ke2017sparse} & $1.49$ & $1.40$ \\ 
 & \textcolor{black}{GRU, BPTT (impl.)} & \textcolor{black}{$1.50$} & \textcolor{black}{$1.41$} \\
 & RNN, BPTT (impl.) & $2.20$ & $2.16$ \\
 & RNN, SAB (impl.) & $2.27$ & $2.19$ \\
 \hline
 \parbox[t]{0.15mm}{\multirow{5}{*}{\rotatebox[origin=c]{90}{no unfold}}} & RHN, KF-RTRL \cite{mujika2018approximating} & $1.77$ & $1.72$ \\ 
 & ESN (impl.) & $3.22$ & $3.16$ \\ 
 & RHN, UORO \cite{tallec2017unbiased} & $2.63$ & $2.61$ \\
 & LSTM, RTRL (impl.) & $1.75$ & $1.71$ \\ 
 & P-TNCN (ours) & $1.73$ & $1.70$ \\ 
 \hline
\end{tabular}
\end{center}
\vspace{-0.45cm} 
\end{table*}

\begin{figure*}
\centering
  \includegraphics[width=0.125\linewidth]{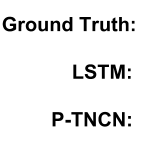}
  ~
  \includegraphics[width=0.75\linewidth]{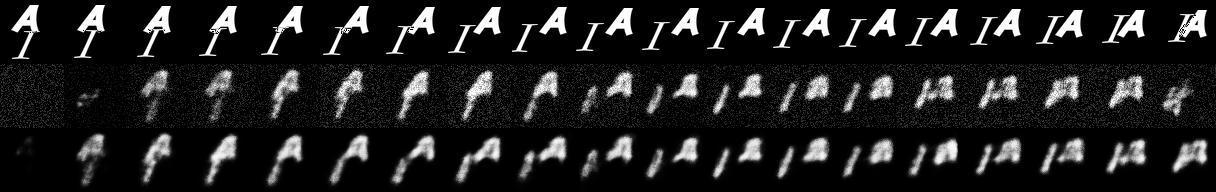}
	\caption{Zero-shot test runs where a model that is trained on \textbf{digits} from Bouncing MNIST and is tested on Bouncing NotMNIST data, patterns it has \textbf{never} before seen (such as \textbf{letters}). The top row contains ground truth frames from a randomly selected test sequence, while the middle row contains the relevant frame by frame predictions made by an LSTM trained with BPTT and the bottom row contains those made by the proposed P-TNCN.}
\label{fig:zero_shot}
\vspace{-0.4cm} 
\end{figure*}

\subsection{Training setup}
We trained P-TNCNs with multiple layers of latent variables, searching for the size of the layers over the range $\{1000-3000\}$ ($3$ layers were used for video data and $2$ layers were used for text/symbol data). The activation $\phi^\ell_z(\cdot)$ was chosen to be the hyperbolic tangent function. Parameters, including the error feedback weights, were initialized from zero-mean Gaussian distributions with $\sigma^2 = 0.025$. The sparsity coefficient was $\lambda = 0.001$, the correction factor was $\beta = 0.15$ and the top-down modulation factor was $\gamma = 0.01$. All meta-parameter values were found via a light, course grid search where generalization performance was measured strictly on each dataset's validation subset.
Parameter updates at each time step were estimated using mini-batches of 20 samples for videos (across 20 parallel sequences), and of 50 samples for text. Parameters were updated using the method of stochastic gradient descent with a step-size of $\lambda = 0.035$, and updates were rescaled (as the gradients were in \cite{pascanu2013difficulty}) to have unit norms. 
Furthermore, we impose a max-norm constraint on the model parameter column vectors\textcolor{black}{. Specifically,} column parameter values were projected, after each update, to the L2 ball, centered at the origin, with radius $l = \{30\}$ (value found in preliminary experiments, though training was not too sensitive to this exact value).
Note that all of our models are given no prior knowledge of the task, e.g. convolutional weight matrices, much as was done in \cite{taylor2007modeling}. 

For RTRL, SAB , BPTT, UORO, and the ESNs, we trained Long Short-Term Memory (LSTM) \cite{hochreiter1997long,greff2017lstmSearch} models with $1000$ units and simple Elman recurrent network (RNN) models with $2000$, tuning meta-parameters by tracking performance on the validation set. We optimize using the RMSprop adaptive learning rate scheme, using a variable learning rate that ranged between $ 0.01$ to $0.0004$ with a batch size of $20$. Training deeper and wider models using RTRL is incredibly computationally expensive, hence we stuck to a standard architecture and tuned the model based on the validation set. On the other hand, UORO required its own separate tuning, since it is a noisy one rank approximation of RTRL, making the optimal meta-parameter choices found for RTRL unusable. For SAB, we obtained better results using $k_{trunc} = 5$, $k_{top} = 10$ and $k_{attn}=2$ (through manual experimentation using validation performance as a guide). 
SAB requires additional hyper-parameters to tune beyond those inherent to BPTT/TBPTT, which are non-trivial to tune given the high computational cost associated with its (jointly-learned) attention mechanism. It is also not clear how to extend such an algorithm to deeper networks and the trade-off between performance and computational cost has yet to be properly analyzed.   
For the ESN models, we perform additional hyper-parameter tuning to adjust their reservoir weights to each given task. This entails a rather time-consuming tuning process, since obtaining ideal hyper-parameter settings for the ESN is also non-trivial.


\subsection{Zero-Shot Adaptation Setup}
\label{sec:zeroshot_setup}
Given that the P-TNCN is engaged in what is essentially a never-ending process of error-correction, even when its weights are not being altered, it would be particularly interesting to investigate the model's ability to process patterns it has clearly never seen before. This setting we will refer to as \emph{zero-shot adaptation}, which is strongly related to the concept of zero-shot learning \cite{kodirov2015unsupervised}. The basic premise of this task is to take a model that is trained on one dataset and apply it to a completely different one, not allowing it to modify its internal synaptic weights. Since the P-TNCN has adaptive behavior built into its very processing, the model might stand a chance at effectively handling novel inputs without additional learning. 
To test zero-shot adaptation, we take the models learned at the end of training on the Bouncing MNIST dataset and apply them to the Bouncing NotMNIST test-set. In addition, we do the same for the \textcolor{black}{reverse and take the models learned on Bouncing NotMNIST and evaluate them on} the Bouncing MNIST test-set.
This would mean a model that has learned the dynamics of moving digits would need to apply part of its knowledge to the dynamics of moving characters/glyphs (and vice versa) in addition to successfully reconstructing such objects, making this task particularly difficult.
\vspace{-0.2cm}

\subsection{Online Continual Learning Setup}
In one final experiment, we investigate a microcosm of the lifelong learning problem \cite{su2019genlifelong,zhang2019hlml}, specifically the case of continual generative modeling, where the model is presented with a series of sequence modeling tasks. In this setting, knowledge retention becomes quite important, given that neural systems are prone to forgetting \cite{mccloskey_catastrophic_1989,fu1996incrementalbp}, and will allow us to investigate if the P-TNCN's error-correction process is a mechanism useful for adapting to novel scenarios while not interfering with its recollection of how it modeled prior tasks.

We start by training models each dataset with only a single pass, with mini-batches of size $1$ (i.e., pure online learning), which is a much more realistic and useful setting when faced with infinite streams of patterns, with data coming from different tasks at different times. The models are one-shot trained on each task's training set in the following order: 1) Bouncing MNIST (task $\mathbf{T}_1$), 2)  Bouncing NotMNIST (task $\mathbf{T}_2$), and 3) Bouncing Fashion MNIST (task $\mathbf{T}_3$). 
\vspace{-0.2cm}

\subsection{Results}
We compare our performance to the baseline algorithms/models described in the last section, e.g., BPTT, SAB, RTRL, UORO, and ESNs. These results include previously reported results when applicable (such as for Kronecker Factored RTRL \cite{mujika2018approximating} approximation of RTRL, KF-RTRL). Note that TBPTT requires unrolling over $T = 25$ steps (TBPTT-$25$). All baselines were tuned on validation for all experiments.

\begin{table*}[t]
\begin{center}
\caption{Task matrices for the P-TNCN and LSTMs trained with BPTT, SAB, or RTRL for continual sequence modeling. Measurements are validation squared error (SE), averaged over $5$ trials. Note: $\mathbf{T}_1$ is MNIST, $\mathbf{T}_2$ is NotMNIST, and $\mathbf{T}_3$ is Fashion MNIST.}
\label{results:continual_results} 
\footnotesize
\begin{tabular}{|c||c|c|c||c|c|c||c|c|c||c|c|c||}
\hline
  & \multicolumn{3}{c||}{\textbf{P-TNCN}} & \multicolumn{3}{c||}{\textbf{LSTM, BPTT}} & \multicolumn{3}{c||}{\textbf{LSTM, SAB}} & \multicolumn{3}{c||}{\textbf{LSTM, RTRL}}\\
  Eval Pt. & $\mathbf{T}_1$ SE & $\mathbf{T}_2$ SE & $\mathbf{T}_3$ SE & $\mathbf{T}_1$ SE & $\mathbf{T}_2$ SE & $\mathbf{T}_3$ SE & $\mathbf{T}_1$ SE & $\mathbf{T}_2$ SE & $\mathbf{T}_3$ SE & $\mathbf{T}_1$ SE & $\mathbf{T}_2$ SE & $\mathbf{T}_3$ SE  \\
  \hline
  After $\mathbf{T}_1$ & 102.67 & -- & -- & 139.89 & -- & -- & 139.27 & -- & -- & 128.89 & -- & -- \\
  After $\mathbf{T}_2$ & 102.79 & 215.65 & -- & 146.97 & 246.59 & -- & 143.59 & 247.81 & -- & 130.12 & 233.60 & -- \\
  After $\mathbf{T}_3$ & 100.31 & 223.28 & 93.36 & 147.85 & 260.62 & 129.91 & 143.99 & 256.27 & 129.88 & 131.01 & 249.63 & 112.99 \\
  \hline
\end{tabular}
\end{center}
\vspace{-0.675cm} 
\end{table*}

\begin{table}[t]
\begin{center}
\caption{Zero-shot adaptive performance of the models trained on NotMNIST and tested on MNIST and vice versa.}
\label{results:adaptation}
\begin{tabular}{ l|ll||ll } 
 \hline 
 & \multicolumn{2}{c||}{\textbf{NotMNIST $\rightarrow$ MNIST}}  & \multicolumn{2}{c}{\textbf{MNIST $\rightarrow$ NotMNIST}} \\
 \hline
 \bf Model, 0-shot & \bf CE & \bf SE & \bf CE & \bf SE\\
 \hline
 \hline
 LSTM, BPTT & $492.21$ & $104.76$ & $1297.26$ & $325.56$ \\
 LSTM, SAB & $499.21$ & $105.87$ & $1299.28$ & $329.59$ \\
 LSTM, RTRL & $447.28$ & $99.89$ & $1211.01$ & $293.56$ \\
 P-TNCN & $\mathbf{377.30}$ & $\mathbf{89.39}$ & $\mathbf{1131.7}$ & $\mathbf{257.07}$ \\
 \hline
\end{tabular}
\end{center}
\vspace{-0.7cm} 
\end{table}

In addition, the continual corrective process of the P-TNCN should also aid it greatly when dealing with out-of-domain inputs/new tasks. We will test this ability in the next experiment, however, in the appendix, we explore this hypothesis qualitatively by examining actual samples of the P-TNCN when processing Bouncing MNIST and NotMNIST videos that vary the number of bouncing objects depicted (following the Bouncing MNIST setup of \cite{srivastava2015video}). 

\subsubsection{Sequence Prediction}
\label{sec:seq_pred}
We report performance for the video and language modeling tasks in Table \ref{results:metrics}. In short, we see that the proposed P-TNCN performs better than all of the online algorithms and models. On both bouncing MNIST and NotMNIST, the P-TNCN outperforms \emph{all} competing approaches, including those based on full BPTT. Note that, for bouncing MNIST, the LSTM models of \cite{srivastava2015video} were trained using a data generator instead of a fixed sample which gives those models an additional advantage when learning better feature maps (in this favorable setting, \cite{srivastava2015video} would not be concerned with overfitting).
Nonetheless, our simple, efficient P-TNCN is still able to outperform those models using only a fixed training set (i.e., no data generator).
In the case of Penn Treebank, the P-TNCN outperforms all other competing online approaches but does not quite yet reach the level of performance of BPTT and SAB, which are approaches that require graph unfolding. The fact that the P-TNCN might not perform quite as strongly in language modeling as it does on video data might stem from the fact that the data is not continuous, but rather discrete-valued. Predictive coding models have almost exclusively been applied to real-valued data and it is not clear if there is some property of discrete/symbolic data that might interact negatively with the process of iterative error correction.
Nonetheless, these results, taken as a whole, provide strong positive evidence that a continual error-correction approach to learning and prediction can serve as an effective alternative to the many various approximations of back-propagation of errors for sequence learning. In short, during the act of processing time-varying data points, the local representation targets that the P-TNCN creates, at each time step, help the model ``stay on track''.

\subsubsection{Zero-Shot Adaptation} 
\label{sec:zeroshot_results}
The results of our zero-shot adaptation experiment are reported in Table \ref{results:adaptation}. Furthermore, the predictions made from the P-TNCN model trained on MNIST are shown as it processes a NotMNIST sample sequence in Figure \ref{fig:zero_shot}.
Promisingly, we observe that the P-TNCN is able to actually perform reasonably well on completely unseen input objects, e.g., able to predict characters when only trained on digits as evidenced by our sampled sequence. Furthermore, again, it generalizes better than RNNs trained with other alternatives, e.g., SAB and RTRL, as well as BPTT.

Qualitatively, we observe in Figure \ref{fig:zero_shot} that, beyond its very noise predictions, the LSTM, nearing the sequence end, starts to blob the characters together and begins to treat the characters as digits. Specifically, it is as if LSTM perceives the $I$ as a $1$ and the $A$ as a $3$, finally blending them into a $4$. The LSTM picks up a basic sense of how the strokes are formed at first, starting to pick up some structure using whatever relevant representation it can find in memory, but as it runs on longer, it forgets, converging to perhaps a mean representation (yielding a crude $4$).
The P-TNCN, while not perfect, is able to retain the idea of two distinct characters, even to the end.
Notably, we observe that SAB struggles with this setup, where it is likely that the attention mechanism itself is the cause (the defining feature of SAB over plain BPTT). Specifically, we hypothesize that while SAB might facilitate improved gradient flow (and a desirable averaging of weights over time) when focused on a single task, it yields distributed feature representations that are too dataset-specific. Rather, SAB restricts how well the model weights can generalize to subsequent, newer datasets (also observed this in the appendix experiment). 

The reason for our model's success in zero-shot adaptation, we hypothesize, is related to the P-TNCN's \textcolor{black}{processing mechanism. Unlike most recurrent network models, the P-TNCN is in a constant state of aggressive error-correction, even weights are not being adapted.} Furthermore, the learning process of the P-TNCN is perhaps facilitating the acquisition of something more general than \textcolor{black}{how to predict objects that occur in a particular sample. In essence, the P-TNCN \emph{is learning how to error correct}, which allows it to deal with out-of-domain inputs.} Future work will entail investigating the full extent of the P-TNCN's ability to dynamically adapt to novel inputs. 

\subsubsection{Online Continual Sequence Modeling} 
\label{sec:continual_results}
Finally, to measure performance in the online continual sequence learning scenario, we present the task performance matrices for each model/algorithm (Table \ref{results:continual_results}). The main diagonal of a matrix contains validation performance (squared error) immediately after training on a task $T_i$. The lower, off-diagonal scores contain the performance on prior tasks, i.e., $\{ T_1, \cdots, T_{i-1} \}$, immediately after training on task $T_i$. Note that in most work on forgetting, classification accuracy is evaluated, while in this work, we focus on generative modeling measurements.

In Table \ref{results:continual_results}, the P-TNCN outperforms the baseline models on each task after one-shot training, i.e., main diagonal measurements (lowest validation error), and its performance on prior modeling tasks does not degrade as much as models trained with BPTT/SAB. Additionally, there appears to be some positive backward transfer from learning tasks $T_2$ and $T_3$ for task $T_1$ which might further confirm the hypothesis that the P-TNCN is learning the process of error-correction. In terms of performance degradation, for task $T_1$, RTRL appears to retain most of its original performance (though always considerably worse than that of the P-TNCN's), but this is not the case for task $T_2$. These results should encourage exploration of the P-TNCN in complex continual learning settings, especially given the fact that the P-TNCN reaches lower generalization error compared to the baselines with only one pass through each task dataset. Since the P-TNCN is dynamic, requires no unrolling like BPTT and SAB, and is computationally cheaper than RTRL, it offers a strong learning algorithm framework for tackling the open challenge of never-ending learning \cite{mitchell2015never}.
\vspace{-0.2cm}

\begin{figure*}
\centering
  \includegraphics[width=0.8\linewidth]{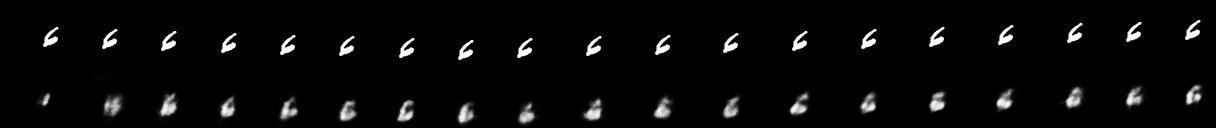}\\
  \vspace{0.075cm}
  \includegraphics[width=0.8\linewidth]{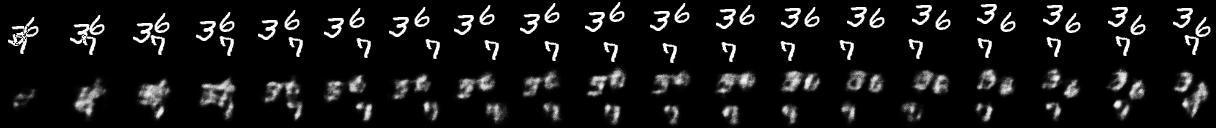}\\
  \vspace{0.075cm}
  \includegraphics[width=0.8\linewidth]{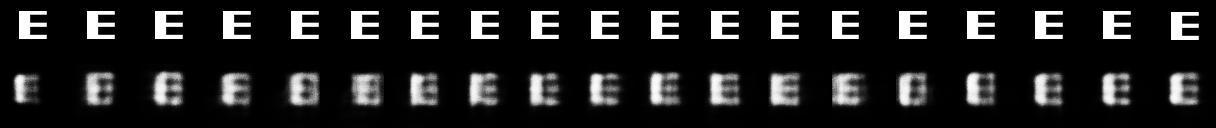}\\
  \vspace{0.075cm}
  \includegraphics[width=0.8\linewidth]{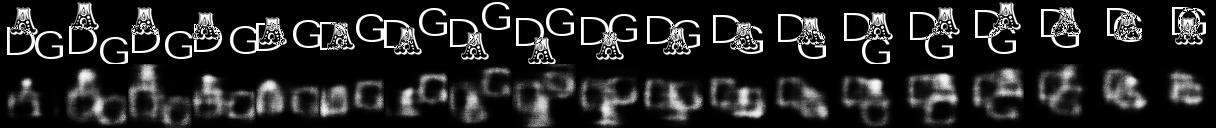}
	\caption{Out-of-domain test runs. Frame by frame predictions (bottom row) of the P-TNCN compared to ground truth (top row) frames on test sequences of one and three moving digits. Recall that the P-TNCN was only trained on sequences of two moving digits.}
\label{fig:samples}
\vspace{-0.5cm} 
\end{figure*}

\section{Conclusion}
We proposed the Parallel Temporal Neural Coding Network, which is a layerwise, parallelizable recurrent neural model, and its learning algorithm based on Local Representation Alignment which can be considered a neurocognitively-plausible alternative to back-propagation of errors. Learning with this architecture does not require any unfolding over data sequences and its underlying architectural design has the potential to exploit the benefits of parallel hardware implementations. In addition, our model does not require differentiable activation functions, which offers opportunities to explore the integration of complex units. Our experiments demonstrate that our model and its learning algorithm outperform key online training algorithms, notably the expensive real-time recurrent learning, and can match or even outperform back-propagation through time. Furthermore, our model is capable of effective zero-shot adaptation and online continual sequence learning.
\vspace{-0.25cm}

\section*{Acknowledgment}
\noindent
The authors gratefully acknowledge partial support from the National Science Foundation.


\appendix  
In this appendix, we present a complexity analysis of the P-TNCN trained via LRA as well as several of the key algorithms used to train RNNs. In addition, we present a qualitative exploration of the P-TNCN's ability to process out-of-domain inputs, an extra experiment in zero-shot adaptation, and a small stream experiment to demonstrate how the model works with discrete-valued nonlinearities.

\begin{table}[t]
\begin{center}
\caption{Complexity analysis of various RNN learning algorithms (for $M$ layers of $n$ units) for both offline and online scenarios.}
\vspace{-0.4cm}
\label{complexity_analysis}
\begin{tabular}{l||ll|ll}
  & \textbf{Offline}  &   & \multicolumn{2}{l}{\textbf{Online (per time step)} } \\
Algorithm & Time  & Space & Time & Space \\
\hline
BPTT & $O(n^2 L M)$ & $O(n^2 M)$ & $O(n^2 T M)$ & $O(n^2 M)$ \\
TBPTT & $O(n^2 h M)$ & $O(n^2 M)$ & $O(n^2 h M)$ & $O(n^2 M)$ \\
RTRL & $O(n^4 M)$ & $O(n^3 M)$ & $O(n^4 M)$ & $O(n^3 M)$  \\
UORO & $O(n^2 L M)$ & $O(n^2 M)$ & $O(n^2 M)$ & $O(n^2 M)$  \\
P-TNCN (LRA) & $O(n^2 L M)$ & $O(n^2 M)$ & $O(n^2 M)$ & $O(n^2 M)$ \\
\hline
\end{tabular}
\end{center}
\vspace{-0.8cm}
\end{table}

\textcolor{black}{
\textbf{On Temporal and Spatial Complexity: }
\label{sec:complexity_analysis}
To appreciate the value of the P-TNCN and its learning procedure, LRA, one should consider the temporal and spatial complexities in both the offline and online stream-driven learning scenarios.  We examine the key algorithms one could use to train an $M$-layered RNN (in this case, the analysis is further restricted to simple Elman-style RNNs), i.e., BPTT, TBPTT, RTRL, and UORO. Each layer has $n$ units. In the \emph{offline} setting, the RNN is to process and adapt sequences of length $L$. In the \emph{online} setting, the RNN is process and adapt to an infinitely-long sequence (i.e. a data stream), where $T$ marks the length of the sequence seen up until the current point of time $t$.
In Table \ref{complexity_analysis} we present the results of this analysis for the settings described above (with some complexity results from \cite{williams1995gradient}, which we extend). Note that (for any time-step) the upper bound calculation for all RNN-based learning approaches is dominated by the computation of the recurrent weight matrix update (an $n \times n$ weight matrix, as was the case in \cite{williams1995gradient}).
}

\textcolor{black}{ 
TBPTT reduces its time complexity, compared to BPTT, in both settings by only unrolling over a history buffer of length $h$. In the \emph{offline} setting, we note that the P-TNCN trained via LRA operates with the same complexity as BPTT. However, note that our approach to training circumvents many of the issues associated with backprop's unrolling, i.e., vanishing gradients and restriction to differentiable activations. P-TNCN has the same time complexity as UORO (but better accuracy, as we shall see in Section \ref{sec:results}) and better time/space complexity than RTRL.
When a stream is being processed step by step, BPTT requires the RNN being trained to be unrolled back until the start of the sequence \emph{at each time step}, yielding a time complexity of $O(n^2 T M)$. TBPTT reduces this a bit by only maintaining a fixed buffer of length $h$, resulting in a complexity of $O(n^2 h M)$. The complexities of P-TNCN (LRA), UORO, and RTRL do not change in the online setting (again, the advantage of P-TNCN here is in accuracy). 
}

\textbf{Out-of-Domain Samples: }
\label{sec:out_of_domain}
To explore the P-TNCN's ability to process out-of-domain inputs, we generated new sequences from the Bouncing MNIST and NotMNIST processes as was done in \cite{srivastava2015video} where each sequence contained either only one object (digit or character) or three objects that bounced around. Since our P-TNCNs were only trained on sequences with two objects bouncing around, they have never been exposed to sequences with one or three objects. In Figure \ref{fig:samples}, we show the predictions generated by a trained P-TNCN and see that it does a reasonably good job at predicting a single object bouncing around, which stands in contrast to what was discovered in \cite{srivastava2015video} (which only investigated the case of Bouncing MNIST), where the LSTM trained with BPTT was found to ``hallucinate'' a second digit over top the first/original one. With respect to the three-object sequence sample, we also observe that the P-TNCN is able to maintain its ability to roughly track multiple objects in space (even if not as easily as it could with two digits), and does \emph{not} seem to merge the digits into blobs as the LSTM of \cite{srivastava2015video} does on MNIST. 
However, even thought the P-TNCN appears to do a much better job of predicting the near future, due to its dynamic error units, incorporating an iterative attention mechanism could serve to further improve the P-TNCN's generative abilities.
\begin{table}[t]
\begin{center}
\caption{One-shot training performance of Bouncing MNIST models on Bouncing NotMNIST followed by zero-shot adaptive performance to Bouncing Fashion MNIST (FMNIST).}
\label{results:oneshot}
\begin{tabular}{ l|ll||ll } 
 \hline 
 & \multicolumn{2}{c||}{\textbf{MNIST $\rightarrow$ NotMNIST}}  & \multicolumn{2}{c}{\textbf{NotMNIST $\rightarrow$ FMNIST}} \\
 \hline
 \bf Model,1-0-shot & \bf CE & \bf SE & \bf CE & \bf SE\\
 \hline
 \hline
 LSTM, BPTT & $882.71$ & $269.65$ & $689.96$ & $101.892$ \\
 LSTM, SAB & $883.61$ & $272.58$ & $691.28$ & $103.778$ \\
 LSTM, RTRL & $863.12$ & $239.62$ & $640.09$ & $97.018$ \\
 P-TNCN & $\mathbf{842.13}$ & $\mathbf{215.01}$ & $\mathbf{603.34}$ & $\mathbf{82.841}$ \\
 \hline
\end{tabular}
\end{center}
\vspace{-0.675cm} 
\end{table}

\textbf{Additional Zero-Shot Adaptation Results: }
This extra experiment extends the zero-shot adaptation experiment presented in the main paper by adding in an intermediate phases of one-shot adaptation/learning before evaluating zero-shot capabilities. This setting could be considered a variation of the zero-shot setting where some aspect of continual generative modeling (of test-sets) has been mixed in. 

Specifically, we first take a model already trained on Bouncing MNIST, after many epochs (dispensing with the extreme, one-shot-only constraint of the continual learning experiment) and dynamically adapt it to the NotMNIST test-set (\emph{one-shot learning}). After this one-shot adaptation phase, we evaluate each model's \emph{zero-shot} adaptivity on Fashion MNIST test-set. 

During one-shot adaptation to NotMNIST, we update parameters using stochastic gradient descent with a fixed step size of $0.01$ and report its test-then-train generalization in Table \ref{results:oneshot}. Finally, after one-shot adaptation to NotMNIST, we report the the zero-shot performance, measured in terms of squared error, on Fashion MNIST. 
The results of this experiment, shown in in Table \ref{results:oneshot}, further demonstrate that the P-TNCN's zero-shot adaptive abilities do not degrade when further learning is permitted. Specifically, even when processing the completely orthogonal Fashion MNIST test-set after one-shot adapting to NotMNIST, the P-TNCN outperforms the LSTM models trained with BPTT, SAB, and RTRL.

\textcolor{black}{
\textbf{Using Discrete-Valued Activation Functions: } One rather interesting property of the P-TNCN is that it does not require knowledge of the first derivative of its pointwise activation functions. This means that one could employ a much wider variety of nonlinearities than one could not use in a standard, differentiable RNN that would require BPTT/SAB/RTRL/UORO-based approaches. As a result, the P-TNCN is more general and thus uniquely suited for a wider variety of applications. We demonstrate that one can indeed train a P-TNCN with nondifferentiable activity, we construct a simple toy problem as our experiment. We fit a $2$-layer ($20$ units in each) model composed of the non-differentiable signum activations to a stream of data generated by the ``noisy cosine function'', or $\mathbf{x}_t = cos(t) + \epsilon$ where $\epsilon \sim \mathcal{N}(0,0.02)$ (we simulate $100K$ discrete steps, where when $k = k + 1$, $t = t + \Delta t$, $\Delta t = 0.05$). We obtain a prequential squared error, or pSE (a test-then-train metric inspired by the online learning literature \cite{gama2013evaluating,ororbia2015online}), of $pSE = 0.0163$ whereas the same exact P-TNCN (but one with differentiable tanh activations) yields a $pSE = 0.0169$. The gold standard model (the cosine function) obtains a $pSE = 0.0006$ and a random/non-adapted baseline model yields $pSE = 1.1057$ (the closer to the gold standard, the better).
}


%



\ifCLASSOPTIONcaptionsoff
  \newpage
\fi



\bibliographystyle{IEEEtran}
\bibliography{lra_rnn}

\begin{thebibliography}{10}
\providecommand{\url}[1]{#1}
\csname url@samestyle\endcsname
\providecommand{\newblock}{\relax}
\providecommand{\bibinfo}[2]{#2}
\providecommand{\BIBentrySTDinterwordspacing}{\spaceskip=0pt\relax}
\providecommand{\BIBentryALTinterwordstretchfactor}{4}
\providecommand{\BIBentryALTinterwordspacing}{\spaceskip=\fontdimen2\font plus
\BIBentryALTinterwordstretchfactor\fontdimen3\font minus
  \fontdimen4\font\relax}
\providecommand{\BIBforeignlanguage}[2]{{%
\expandafter\ifx\csname l@#1\endcsname\relax
\typeout{** WARNING: IEEEtran.bst: No hyphenation pattern has been}%
\typeout{** loaded for the language `#1'. Using the pattern for}%
\typeout{** the default language instead.}%
\else
\language=\csname l@#1\endcsname
\fi
#2}}
\providecommand{\BIBdecl}{\relax}
\BIBdecl

\bibitem{stauffer2000learning}
C.~Stauffer and W.~E.~L. Grimson, ``Learning patterns of activity using
  real-time tracking,'' \emph{IEEE Transactions on pattern analysis and machine
  intelligence}, vol.~22, no.~8, pp. 747--757, 2000.

\bibitem{poppe2007vision}
R.~Poppe, ``Vision-based human motion analysis: An overview,'' \emph{Computer
  vision and image understanding}, vol. 108, no. 1-2, pp. 4--18, 2007.

\bibitem{prabhakar2010temporal}
K.~Prabhakar, S.~Oh, P.~Wang, G.~D. Abowd, and J.~M. Rehg, ``Temporal causality
  for the analysis of visual events,'' in \emph{Computer Vision and Pattern
  Recognition (CVPR), 2010 IEEE Conference on}.\hskip 1em plus 0.5em minus
  0.4em\relax IEEE, 2010, pp. 1967--1974.

\bibitem{winograd1972understanding}
T.~Winograd, ``Understanding natural language,'' \emph{Cognitive psychology},
  vol.~3, no.~1, pp. 1--191, 1972.

\bibitem{walsh2010integrating}
T.~J. Walsh, S.~Goschin, and M.~L. Littman, ``Integrating sample-based planning
  and model-based reinforcement learning.'' in \emph{AAAI}, 2010.

\bibitem{bahdanau2014neural}
D.~Bahdanau, K.~Cho, and Y.~Bengio, ``Neural machine translation by jointly
  learning to align and translate,'' \emph{arXiv preprint arXiv:1409.0473},
  2014.

\bibitem{mikolov2010recurrent}
T.~Mikolov, M.~Karafi{\'a}t, L.~Burget, J.~Cernock{\`y}, and S.~Khudanpur,
  ``Recurrent neural network based language model.'' in \emph{Interspeech},
  vol.~2, 2010, p.~3.

\bibitem{ororbia2017diff}
\BIBentryALTinterwordspacing
A.~G. Ororbia, T.~Mikolov, and D.~Reitter, ``Learning simpler language models
  with the differential state framework,'' \emph{Neural Computation}, vol.~0,
  no.~0, pp. 1--26, 2017, pMID: 28957029. [Online]. Available:
  \url{https://doi.org/10.1162/neco\_a\_01017}
\BIBentrySTDinterwordspacing

\bibitem{serban2017piecewise}
I.~V. Serban, A.~G. Ororbia, J.~Pineau, and A.~Courville, ``Piecewise latent
  variables for neural variational text processing,'' in \emph{Proceedings of
  the 2017 Conference on Empirical Methods in Natural Language Processing},
  2017, pp. 422--432.

\bibitem{gopalakrishnan2018neural}
A.~Gopalakrishnan, A.~Mali, D.~Kifer, C.~L. Giles, and A.~G. Ororbia, ``A
  neural temporal model for human motion prediction,'' \emph{arXiv preprint
  arXiv:1809.03036}, 2018.

\bibitem{graves2013speech}
A.~Graves, A.-r. Mohamed, and G.~Hinton, ``Speech recognition with deep
  recurrent neural networks,'' in \emph{Acoustics, speech and signal processing
  (icassp), 2013 IEEE international conference on}.\hskip 1em plus 0.5em minus
  0.4em\relax IEEE, 2013, pp. 6645--6649.

\bibitem{ororbia2017learning}
A.~G. Ororbia, P.~Haffner, D.~Reitter, and C.~L. Giles, ``Learning to adapt by
  minimizing discrepancy,'' \emph{arXiv preprint arXiv:1711.11542}, 2017.

\bibitem{rumelhart1988backprop}
\BIBentryALTinterwordspacing
D.~E. Rumelhart, G.~E. Hinton, and R.~J. Williams, ``Neurocomputing:
  Foundations of research,'' J.~A. Anderson and E.~Rosenfeld, Eds.\hskip 1em
  plus 0.5em minus 0.4em\relax Cambridge, MA, USA: MIT Press, 1988, ch.
  Learning Representations by Back-propagating Errors, pp. 696--699. [Online].
  Available: \url{http://dl.acm.org/citation.cfm?id=65669.104451}
\BIBentrySTDinterwordspacing

\bibitem{ororbia2018conducting}
A.~G. Ororbia, A.~Mali, D.~Kifer, and C.~L. Giles, ``Deep credit assignment by
  aligning local representations,'' \emph{arXiv preprint arXiv:1803.01834},
  2018.

\bibitem{ororbia2018biologically}
A.~G. Ororbia and A.~Mali, ``Biologically motivated algorithms for propagating
  local target representations,'' \emph{arXiv preprint arXiv:1805.11703}, 2018.

\bibitem{rao1999predictive}
R.~P. Rao and D.~H. Ballard, ``Predictive coding in the visual cortex: a
  functional interpretation of some extra-classical receptive-field effects.''
  \emph{Nature neuroscience}, vol.~2, no.~1, 1999.

\bibitem{panichello2013predictive}
\BIBentryALTinterwordspacing
M.~Panichello, O.~Cheung, and M.~Bar, ``Predictive feedback and conscious
  visual experience,'' \emph{Frontiers in Psychology}, vol.~3, p. 620, 2013.
  [Online]. Available:
  \url{https://www.frontiersin.org/article/10.3389/fpsyg.2012.00620}
\BIBentrySTDinterwordspacing

\bibitem{rao1997dynamic}
R.~P. Rao and D.~H. Ballard, ``Dynamic model of visual recognition predicts
  neural response properties in the visual cortex,'' \emph{Neural computation},
  vol.~9, no.~4, pp. 721--763, 1997.

\bibitem{lillicrap2014random}
T.~P. Lillicrap, D.~Cownden, D.~B. Tweed, and C.~J. Akerman, ``Random feedback
  weights support learning in deep neural networks,'' \emph{arXiv preprint
  arXiv:1411.0247}, 2014.

\bibitem{lee2015difference}
D.-H. Lee, S.~Zhang, A.~Fischer, and Y.~Bengio, ``Difference target
  propagation,'' in \emph{Joint European Conference on Machine Learning and
  Knowledge Discovery in Databases}.\hskip 1em plus 0.5em minus 0.4em\relax
  Springer, 2015, pp. 498--515.

\bibitem{scellier2017equilibrium}
B.~Scellier and Y.~Bengio, ``Equilibrium propagation: Bridging the gap between
  energy-based models and backpropagation,'' \emph{Frontiers in computational
  neuroscience}, vol.~11, p.~24, 2017.

\bibitem{taylor2007modeling}
G.~W. Taylor, G.~E. Hinton, and S.~T. Roweis, ``Modeling human motion using
  binary latent variables,'' in \emph{Advances in neural information processing
  systems}, 2007, pp. 1345--1352.

\bibitem{williams1989learning}
R.~J. Williams and D.~Zipser, ``A learning algorithm for continually running
  fully recurrent neural networks,'' \emph{Neural computation}, vol.~1, no.~2,
  pp. 270--280, 1989.

\bibitem{ollivier2015training}
Y.~Ollivier, C.~Tallec, and G.~Charpiat, ``Training recurrent networks online
  without backtracking,'' \emph{arXiv preprint arXiv:1507.07680}, 2015.

\bibitem{tallec2017unbiased}
C.~Tallec and Y.~Ollivier, ``Unbiased online recurrent optimization,''
  \emph{arXiv preprint arXiv:1702.05043}, 2017.

\bibitem{mujika2018approximating}
A.~Mujika, F.~Meier, and A.~Steger, ``Approximating real-time recurrent
  learning with random kronecker factors,'' \emph{arXiv preprint
  arXiv:1805.10842}, 2018.

\bibitem{ke2017sparse}
N.~R. Ke, A.~G., O.~Bilaniuk, J.~Binas, L.~Charlin, C.~Pal, and Y.~Bengio,
  ``Sparse attentive backtracking: Long-range credit assignment in recurrent
  networks,'' \emph{arXiv preprint arXiv:1711.02326}, 2017.

\bibitem{sutskever2009recurrent}
I.~Sutskever, G.~E. Hinton, and G.~W. Taylor, ``The recurrent temporal
  restricted boltzmann machine,'' in \emph{Advances in Neural Information
  Processing Systems}, 2009, pp. 1601--1608.

\bibitem{jaeger2002tutorial}
H.~Jaeger, \emph{Tutorial on training recurrent neural networks, covering BPPT,
  RTRL, EKF and the" echo state network" approach}.\hskip 1em plus 0.5em minus
  0.4em\relax GMD-Forschungszentrum Informationstechnik Bonn, 2002, vol.~5.

\bibitem{maass2004lsm}
\BIBentryALTinterwordspacing
W.~Maass and H.~Markram, ``On the computational power of circuits of spiking
  neurons,'' \emph{Journal of Computer and System Sciences}, vol.~69, no.~4,
  pp. 593 -- 616, 2004. [Online]. Available:
  \url{http://www.sciencedirect.com/science/article/pii/S0022000004000406}
\BIBentrySTDinterwordspacing

\bibitem{schiller2005analyzing}
U.~D. Schiller and J.~J. Steil, ``Analyzing the weight dynamics of recurrent
  learning algorithms,'' \emph{Neurocomputing}, vol.~63, pp. 5--23, 2005.

\bibitem{rainer1999prospective}
G.~Rainer, S.~C. Rao, and E.~K. Miller, ``Prospective coding for objects in
  primate prefrontal cortex,'' \emph{Journal of Neuroscience}, vol.~19, no.~13,
  pp. 5493--5505, 1999.

\bibitem{olshausen1997sparse}
B.~A. Olshausen and D.~J. Field, ``Sparse coding with an overcomplete basis
  set: A strategy employed by v1?'' \emph{Vision research}, vol.~37, no.~23,
  pp. 3311--3325, 1997.

\bibitem{rauss2013predictive}
\BIBentryALTinterwordspacing
K.~Rauss and G.~Pourtois, ``What is bottom-up and what is top-down in
  predictive coding?'' \emph{Frontiers in Psychology}, vol.~4, p. 276, 2013.
  [Online]. Available:
  \url{https://www.frontiersin.org/article/10.3389/fpsyg.2013.00276}
\BIBentrySTDinterwordspacing

\bibitem{chalasani2013deep}
R.~Chalasani and J.~C. Principe, ``Deep predictive coding networks,''
  \emph{arXiv preprint arXiv:1301.3541}, 2013.

\bibitem{chalasani2015contextcdn}
R.~{Chalasani} and J.~C. {Principe}, ``Context dependent encoding using
  convolutional dynamic networks,'' \emph{IEEE Transactions on Neural Networks
  and Learning Systems}, vol.~26, no.~9, pp. 1992--2004, Sep. 2015.

\bibitem{lotter2016deep}
W.~Lotter, G.~Kreiman, and D.~Cox, ``Deep predictive coding networks for video
  prediction and unsupervised learning,'' \emph{arXiv preprint
  arXiv:1605.08104}, 2016.

\bibitem{santana2017exploiting}
E.~Santana, M.~S. Emigh, P.~Zegers, and J.~C. Principe, ``Exploiting
  spatio-temporal structure with recurrent winner-take-all networks,''
  \emph{IEEE Transactions on Neural Networks and Learning Systems}, 2017.

\bibitem{hwang2018robotPC}
J.~{Hwang}, J.~{Kim}, A.~{Ahmadi}, M.~{Choi}, and J.~{Tani}, ``Dealing with
  large-scale spatio-temporal patterns in imitative interaction between a robot
  and a human by using the predictive coding framework,'' \emph{IEEE
  Transactions on Systems, Man, and Cybernetics: Systems}, pp. 1--14, 2018.

\bibitem{song2018fastinfpc}
Z.~{Song}, J.~{Zhang}, G.~{Shi}, and J.~{Liu}, ``Fast inference predictive
  coding: A novel model for constructing deep neural networks,'' \emph{IEEE
  Transactions on Neural Networks and Learning Systems}, vol.~30, no.~4, pp.
  1150--1165, April 2019.

\bibitem{balduzzi2015kickback}
D.~Balduzzi, H.~Vanchinathan, and J.~M. Buhmann, ``Kickback cuts backprop's
  red-tape: Biologically plausible credit assignment in neural networks.'' in
  \emph{AAAI}, 2015, pp. 485--491.

\bibitem{jaderberg2016decoupled}
M.~Jaderberg, W.~M. Czarnecki, S.~Osindero, O.~Vinyals, A.~Graves, and
  K.~Kavukcuoglu, ``Decoupled neural interfaces using synthetic gradients,''
  \emph{arXiv preprint arXiv:1608.05343}, 2016.

\bibitem{nokland2016direct}
A.~N{\o}kland, ``Direct feedback alignment provides learning in deep neural
  networks,'' in \emph{Advances in Neural Information Processing Systems},
  2016, pp. 1037--1045.

\bibitem{ororbia_deep_hybrid_2015a}
A.~G. Ororbia, D.~Reitter, J.~Wu, and C.~L. Giles, ``Online learning of deep
  hybrid architectures for semi-supervised categorization,'' in \emph{Machine
  {Learning} and {Knowledge} {Discovery} in {Databases} ({Proceedings}, {ECML}
  {PKDD} 2015)}, ser. Lecture {Notes} in {Computer} {Science}.\hskip 1em plus
  0.5em minus 0.4em\relax Porto, Portugal: Springer, 2015, vol. 9284, pp.
  516--532.

\bibitem{bengio_greedy_2007}
Y.~Bengio, P.~Lamblin, D.~Popovici, H.~Larochelle \emph{et~al.}, ``Greedy
  layer-wise training of deep networks,'' \emph{Advances in neural information
  processing systems}, vol.~19, p. 153, 2007.

\bibitem{hopfield1982neural}
J.~J. Hopfield, ``Neural networks and physical systems with emergent collective
  computational abilities,'' \emph{Proceedings of the national academy of
  sciences}, vol.~79, no.~8, pp. 2554--2558, 1982.

\bibitem{battaglia2004local}
F.~P. Battaglia, G.~R. Sutherland, and B.~L. McNaughton, ``Local sensory cues
  and place cell directionality: additional evidence of prospective coding in
  the hippocampus,'' \emph{Journal of Neuroscience}, vol.~24, no.~19, pp.
  4541--4550, 2004.

\bibitem{schutz2007prospective}
S.~Sch{\"u}tz-Bosbach and W.~Prinz, ``Prospective coding in event
  representation,'' \emph{Cognitive processing}, vol.~8, no.~2, pp. 93--102,
  2007.

\bibitem{o1996biologically}
R.~C. O'Reilly, ``Biologically plausible error-driven learning using local
  activation differences: The generalized recirculation algorithm,''
  \emph{Neural computation}, vol.~8, no.~5, pp. 895--938, 1996.

\bibitem{oreilly1998sixprinciples}
------, ``Six principles for biologically based computational models of
  cortical cognition,'' \emph{Trends in cognitive sciences}, vol.~2, no.~11,
  pp. 455--462, 1998.

\bibitem{druck2007semi}
G.~Druck, C.~Pal, A.~McCallum, and X.~Zhu, ``Semi-supervised classification
  with hybrid generative/discriminative methods,'' in \emph{Proceedings of the
  13th ACM SIGKDD international conference on Knowledge discovery and data
  mining}.\hskip 1em plus 0.5em minus 0.4em\relax ACM, 2007, pp. 280--289.

\bibitem{williams1989experimental}
R.~J. Williams and D.~Zipser, ``Experimental analysis of the real-time
  recurrent learning algorithm,'' \emph{Connection Science}, vol.~1, no.~1, pp.
  87--111, 1989.

\bibitem{williams1995gradient}
------, ``Gradient-based learning algorithms for recurrent networks and their
  computational complexity,'' \emph{Backpropagation: Theory, architectures, and
  applications}, vol. 433, 1995.

\bibitem{werbos1988generalization}
P.~J. Werbos, ``Generalization of backpropagation with application to a
  recurrent gas market model,'' \emph{Neural networks}, vol.~1, no.~4, pp.
  339--356, 1988.

\bibitem{werbos1990backpropagation}
------, ``Backpropagation through time: what it does and how to do it,''
  \emph{Proceedings of the IEEE}, vol.~78, no.~10, pp. 1550--1560, 1990.

\bibitem{srivastava2015video}
\BIBentryALTinterwordspacing
N.~Srivastava, E.~Mansimov, and R.~Salakhutdinov, ``Unsupervised learning of
  video representations using lstms,'' in \emph{Proceedings of the 32Nd
  International Conference on International Conference on Machine Learning -
  Volume 37}, ser. ICML'15.\hskip 1em plus 0.5em minus 0.4em\relax JMLR.org,
  2015, pp. 843--852. [Online]. Available:
  \url{http://dl.acm.org/citation.cfm?id=3045118.3045209}
\BIBentrySTDinterwordspacing

\bibitem{xiao2017fashion}
H.~Xiao, K.~Rasul, and R.~Vollgraf, ``Fashion-mnist: a novel image dataset for
  benchmarking machine learning algorithms,'' \emph{arXiv preprint
  arXiv:1708.07747}, 2017.

\bibitem{marcus1993building}
M.~P. Marcus, M.~A. Marcinkiewicz, and B.~Santorini, ``Building a large
  annotated corpus of {English}: The {Penn Treebank},'' \emph{Computational
  Linguistics}, vol.~19, no.~2, pp. 313--330, 1993.

\bibitem{pascanu2013difficulty}
R.~Pascanu, T.~Mikolov, and Y.~Bengio, ``On the difficulty of training
  recurrent neural networks,'' in \emph{International Conference on Machine
  Learning}, 2013, pp. 1310--1318.

\bibitem{hochreiter1997long}
S.~Hochreiter and J.~Schmidhuber, ``Long short-term memory,'' \emph{Neural
  computation}, vol.~9, no.~8, pp. 1735--1780, 1997.

\bibitem{greff2017lstmSearch}
K.~{Greff}, R.~K. {Srivastava}, J.~{Koutník}, B.~R. {Steunebrink}, and
  J.~{Schmidhuber}, ``Lstm: A search space odyssey,'' \emph{IEEE Transactions
  on Neural Networks and Learning Systems}, vol.~28, no.~10, pp. 2222--2232,
  Oct 2017.

\bibitem{kodirov2015unsupervised}
E.~Kodirov, T.~Xiang, Z.~Fu, and S.~Gong, ``Unsupervised domain adaptation for
  zero-shot learning,'' in \emph{Proceedings of the IEEE International
  Conference on Computer Vision}, 2015, pp. 2452--2460.

\bibitem{su2019genlifelong}
X.~{Su}, S.~{Guo}, T.~{Tan}, and F.~{Chen}, ``Generative memory for lifelong
  learning,'' \emph{IEEE Transactions on Neural Networks and Learning Systems},
  pp. 1--15, 2019.

\bibitem{zhang2019hlml}
T.~{Zhang}, G.~{Su}, C.~{Qing}, X.~{Xu}, B.~{Cai}, and X.~{Xing},
  ``Hierarchical lifelong learning by sharing representations and integrating
  hypothesis,'' \emph{IEEE Transactions on Systems, Man, and Cybernetics:
  Systems}, pp. 1--11, 2019.

\bibitem{mccloskey_catastrophic_1989}
M.~{McCloskey} and N.~J. Cohen, ``Catastrophic interference in connectionist
  networks: The sequential learning problem,'' \emph{The psychology of learning
  and motivation}, vol.~24, no. 109, p.~92, 1989.

\bibitem{fu1996incrementalbp}
{LiMin Fu}, {Hui-Huang Hsu}, and J.~C. {Principe}, ``Incremental
  backpropagation learning networks,'' \emph{IEEE Transactions on Neural
  Networks}, vol.~7, no.~3, pp. 757--761, May 1996.

\bibitem{mitchell2015never}
T.~M. Mitchell, W.~W. Cohen, E.~R. Hruschka~Jr, P.~P. Talukdar, J.~Betteridge,
  A.~Carlson, B.~D. Mishra, M.~Gardner, B.~Kisiel, J.~Krishnamurthy
  \emph{et~al.}, ``Never ending learning.'' in \emph{AAAI}, 2015, pp.
  2302--2310.

\bibitem{gama2013evaluating}
J.~Gama, R.~Sebasti{\~a}o, and P.~P. Rodrigues, ``On evaluating stream learning
  algorithms,'' \emph{Machine learning}, vol.~90, no.~3, pp. 317--346, 2013.

\bibitem{ororbia2015online}
A.~G. Ororbia, C.~L. Giles, and D.~Reitter, ``Online semi-supervised learning
  with deep hybrid boltzmann machines and denoising autoencoders,'' \emph{arXiv
  preprint arXiv:1511.06964}, 2015.

\end{thebibliography}

\end{document}